\documentclass[runningheads]{llncs}

\usepackage{eccv}

\usepackage{colortbl}
\usepackage{multirow}

\usepackage{eccvabbrv}

\usepackage{graphicx}
\usepackage{booktabs}

\usepackage[accsupp]{axessibility}  %

\usepackage[breaklinks]{hyperref}

\makeatletter
\g@addto@macro{\endtabular}{\rowfont{}}%
\makeatother
\newcommand{\rowfonttype}{}%
\newcommand{\rowfont}[1]{%
\gdef\rowfonttype{#1}#1\ignorespaces%
}
\makeatother

\definecolor{almond}{rgb}{0.94, 0.87, 0.8}

\newcommand{\fix}[1]{{#1}}
\newcommand{\methodname}{HowToCaption}
\newcommand{\datasetname}{HowToCaption}

\newcommand{\myparagraph}[1]{\vspace{3pt}\noindent{\bf #1}}

\begin{document}

\title{HowToCaption: Prompting LLMs to Transform Video Annotations at Scale}

\author{
Nina Shvetsova\thanks{Equal contribution.}$^{1,2,3}$, Anna Kukleva$^{*2}$, Xudong Hong$^{2,6}$,\\\, Christian Rupprecht$^{4}$, Bernt Schiele$^{2}$, Hilde Kuehne$^{1,3,5}$ \\
}

\authorrunning{N.~Shvetsova et al.}

\institute{$^{1}$Goethe University Frankfurt,
\small $^{2}$MPI for Informatics, SIC, \small $^{3}$University of Bonn, \\ $^{4}$University of Oxford, \small $^{5}$MIT-IBM Watson AI Lab, $^{6}$Saarland University \\
\email{{\{nshvetso,akukleva\}@mpi-inf.mpg.de}}}

\maketitle

\begin{sloppypar} 

\begin{abstract}

Instructional videos are a common source for learning text-video or even multimodal representations by leveraging subtitles extracted with automatic speech recognition systems (ASR) from the audio signal in the videos. However, in contrast to human-annotated captions, both speech and subtitles naturally differ from the visual content of the videos and thus provide only noisy supervision. As a result, large-scale annotation-free web video training data remains sub-optimal for training text-video models. In this work, we propose to leverage the capabilities of large language models (LLMs) to obtain high-quality video descriptions aligned with videos at scale. Specifically, we prompt an LLM to create plausible video captions based on ASR subtitles of instructional videos. To this end, we introduce a prompting method that is able to take into account a longer text of subtitles, allowing us to capture the contextual information beyond one single sentence. We further prompt the LLM to generate timestamps for each produced caption based on the timestamps of the subtitles and finally align the generated captions to the video temporally. In this way, we obtain human-style video captions at scale without human supervision. We apply our method to the subtitles of the HowTo100M dataset, creating a new large-scale dataset, HowToCaption. Our evaluation shows that the resulting captions not only significantly improve the performance over many different benchmark datasets for zero-shot text-video retrieval and video captioning, but also lead to a disentangling of textual narration from the audio, boosting the performance in text-video-audio tasks.\footnote{All data and code is available at \url{https://github.com/ninatu/howtocaption}. }


\end{abstract}

\section{Introduction}

\begin{figure*}[]
\begin{center}
\includegraphics[width=0.9\linewidth]{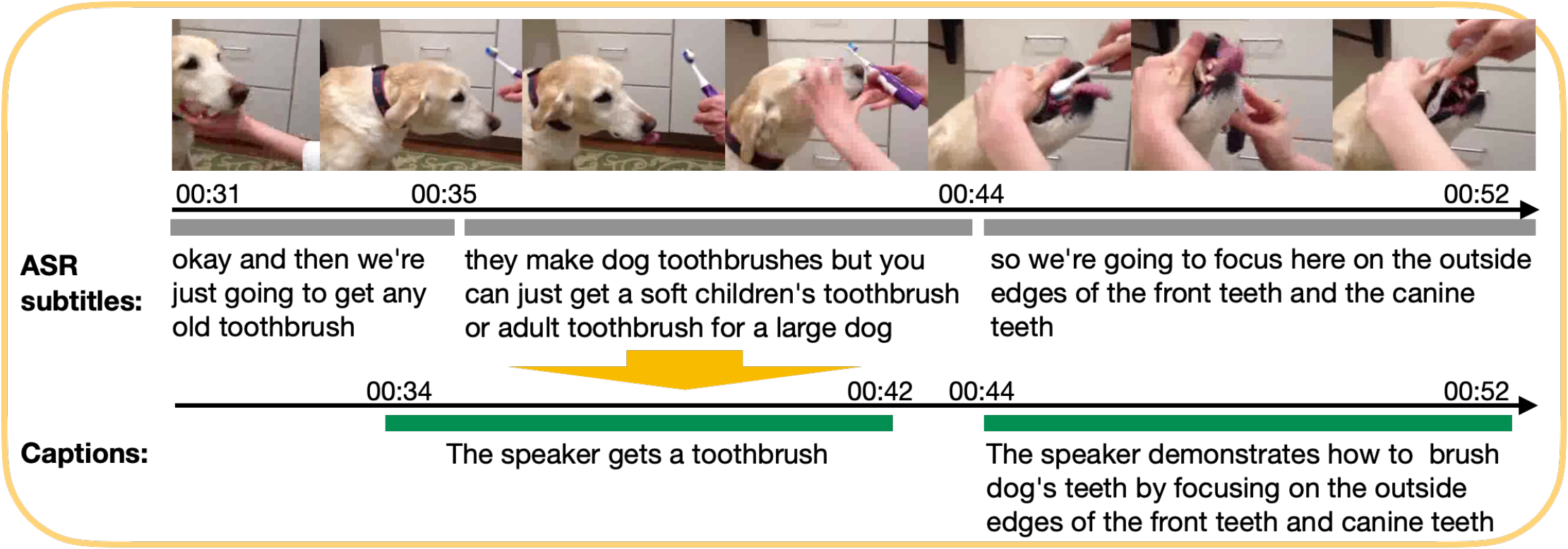}
\end{center}
\caption{ASR subtitles deviate from human-written captions: they contain a lot of filler phrases, e.g., ``we're going to'', and extra information, e.g., ``they make dog toothbrushes''. We propose to generate human-style video captions based on the ASR subtitles and their timestamps that we further temporally realign with the video. 
\label{fig:teaser2}
}
\end{figure*}

Textual descriptions of visual information allow for navigating large amounts of visual data. 
Recently, image-text cross-modal learning has achieved remarkable performance in many downstream tasks by pre-training on large-scale web datasets consisting of text-image pairs~\cite{radford21clip,schuhmann2021laion,jia2021scaling}. 
To collect video data on a similar scale, media platforms such as YouTube can be used as a great source of freely available videos~\cite{abu2016youtube, zhou2018towards,stroud2020learning,miech2019howto100m,zellers2021merlot,xue2022advancing}. 
Most of these videos include some narrations, \eg, in instructional videos~\cite{miech2019howto100m}, people explain and show how to accomplish one or another task. To transform spoken language from the videos into subtitles, current automatic speech recognition (ASR) systems~\cite{radford2023robust} can be used, 
providing aligned text-video annotated pairs for free. 
This automatic supervisory signal can easily scale to large video datasets. 
However, such video web data poses additional challenges~\cite{han2022temporal, miech2019howto100m}: (1) spoken and visual information in the video can deviate from each other, \eg, when speakers provide information beyond what is visible or when spoken instructions do not temporally align with the actions shown, (2) speech contains filler words and phrases, such as ``I'm going to'', and can be incomplete and 
 contain 
grammatical errors, and (3) ASR transcripts usually do not have punctuation and may contain errors (\cref{fig:teaser2}). Therefore, ASR subtitles provide only weak, noisy supervision for videos. %

To address this problem, we propose a new framework, \methodname, that leverages large language models (LLMs)~\cite{vicuna2023} to generate human-style captions on a large scale for web-video instructional datasets based on corresponding ASR subtitles (\cref{fig:teaser2}). By carefully designing prompts, we show that the LLM can effectively map long, noisy subtitles into concise and descriptive human-style video captions. Moreover, we obtain an initial temporal alignment of the generated captions to the video based on the ASR timestamps by tasking the LLM to predict timestamps for each caption. For additional quality improvement, we apply alignment and filtering within short temporal windows with respect to the predicted timestamp. This approach can generate aligned text-video pairs on a large scale without any human intervention.

Beyond providing better annotation, the new captions provide the advantage that they are no longer a direct output of the speech signal, thus effectively decoupling audio and text. Current methods usually avoid using audio~\cite{miech2020end,han2022temporal}, as the ASR subtitles are directly derived from the speech, thus leading to the problem that any text-to-audio+video retrieval would mainly retrieve the closest speech signal while disregarding the video~\cite{shvetsova2022everything,nagrani2022learning}. Being able to generate captions that deviate from the speech thus allows to extend retrieval to audio+video without the need for fine-tuned regularization, as used in~\cite{shvetsova2022everything}.

To verify the effectiveness of the proposed \methodname\ method, we generate new captions for the large-scale HowTo100M dataset~\cite{miech2019howto100m}, obtaining a new \textit{HowToCaption} dataset. We evaluate the quality of the improved textual descriptions on various challenging downstream tasks over four different datasets, namely YouCook2~\cite{zhou2018towards}, MSR-VTT~\cite{xu2016msr}, MSVD~\cite{chen2011collecting}, and LSMDC~\cite{rohrbach2015long}. It shows that the generated captions not only provide a better training signal but also allow for a decoupling of speech and caption annotation, allowing a retrieval based on audio, vision, and subtitles at scale. We release the new \datasetname\ dataset with high-quality textual descriptions to show the potential of generated captions for web text-video pairs. We summarize the contributions of the paper as follows: 
\begin{itemize}
    \item We propose a \methodname\ method to efficiently convert noisy ASR subtitles of instructional videos into accurate video captions, which leverages recent advances in LLMs and generates high-quality video captions at scale without any human supervision. 
    
    \item We create a new \datasetname\ dataset with high-quality human-style textual descriptions with our proposed \methodname\ method.

    \item Utilizing the \datasetname\ dataset for training text-video models allows us to significantly improve the performance over many benchmarks for text-to-video retrieval \fix{and video captioning}. Moreover, since the new textual annotation allows us to disentangle audio and language modalities in the instructional videos, where the ASR subtitles were highly correlated to audio, we show a boost in text-video+audio retrieval performance.

\end{itemize}

\section{Related Work}

\subsection{Large-Scale Video-Language Datasets}

Manual annotation of video captioning datasets is 
extremely
time-consuming 
since it involves video trimming and localization of caption boundaries. 
Currently, manually annotated datasets, e.g., MSR-VTT~\cite{xu2016msr},  YouCook2~\cite{zhou2018towards}, HIREST~\cite{zala2023hierarchical}, and HT-Step~\cite{afouras2024ht}, are limited in size.
Therefore, different methods of mining videos with weak supervision from the Internet were considered. Datasets such as YouTube-8M~\cite{abu2016youtube} and IG-Kinetics-65M~\cite{ghadiyaram2019large} provided multiple class labels based on query clicks, metadata~\cite{abu2016youtube} or hashtags~\cite{ghadiyaram2019large}. 
However, short class labels are suboptimal supervision compared to textual descriptions~\cite{desai2021virtex}. Therefore, Bain et al.~\cite{Bain21} considered scrapping videos with associated alt-text from the web, obtaining the WebVid2M and WebVid10M datasets~\cite{Bain21} with 2M and 10M video-text pairs, respectively.
Stroud et al.~\cite{stroud2020learning} 
proposed to use meta information, such as titles, descriptions, and tags from YouTube, as a textual annotation and created the WTS-70M dataset. Nagrani et al.~\cite{nagrani2022learning} proposed to transfer image captions from an image-text dataset to videos by searching videos with similar frames to an image and collected the VideoCC3M dataset.
Yang et al.~\cite{yang2024vidchapters} created the VidChapters-7M dataset by scraping user-annotated chapters on YouTube. 
However, most videos in WebVid do not have audio, which is an essential part of video analysis, and captions in VideoCC3M are derived from images and, therefore, tend to describe more static scenes rather than actions. At the same time, the title, tags, and chapters of WTS-70M and VidChapters-7M provide only high-level video descriptions. 

As an alternative to this, Miech et al.~\cite{miech2019howto100m} proposed the HowTo100M dataset, where instructional videos are naturally accompanied by dense textual supervision in the form of subtitles obtained from ASR (Automatic Speech Recognition) systems. The HowTo100M dataset with 137M clips sourced from 1.2M YouTube videos was proven to be effective for pre-training video-audio-language representations~\cite{rouditchenko2020avlnet,chen2021multimodal,shvetsova2022everything}.
The followed-up YT-Temporal-180M~\cite{zellers2021merlot} and HD-VILA-100M~\cite{xue2022advancing} datasets follow the same idea but contain more videos with higher diversity and higher video resolution.
While ASR supervision provides a scalable way to create large datasets with dense annotation, the quality of subtitles is not on par with human-annotated captions. In this work, we propose a method to create high-quality captions for videos at scale by leveraging LLM and subtitles.

Recent works, InternVid~\cite{wang2023internvid}, VAST~\cite{chen2024vast}, and Video-ChatGPT~\cite{maaz2023video},  
collect large-scale video-text datasets through per-frame image captioning and text summarization with LLMs.
In contrast, our work stands apart in its objective. Rather than distilling existing knowledge from image models, pre-trained on human-annotated image-text data, into videos, we aim to gather a dataset enriched with new video knowledge. This is achieved by converting freely available ASR subtitles into captions.
Concurrent work, HowToStep~\cite{li2023strong}, summarizes ASR subtitles into descriptive steps using LLM and then performs additional temporal realignment. Our main difference is that we focus on general video captions rather than procedural steps.

\subsection{Learning with Noisy ASR Subtitles of Instructional Videos}

The problem of misalignment and noisiness of ASR supervision in instructional videos, e.g., in the HowTo100M dataset, were addressed in multiple works. MIL-NCE loss~\cite{miech2020end} and soft max-margin ranking loss~\cite{amrani2021noise} were proposed to adapt contrastive loss to misalignment in text-video pairs. Zellers et al.~\cite{zellers2021merlot} proposed to use LLM to add punctuation and capitalization to ASR subtitles and remove mistranscription errors. Han et al.~\cite{han2022temporal} proposed to train temporal alignment networks to filter out subtitles that are not alignable to the video and determine alignment for the others. However, to the best of our knowledge, \cite{lin2022learning}~is the only work that goes beyond just removing mistranscription errors and ASR re-alignment, where Lin et al. proposed to match subtitles to step descriptions from WikiHow dataset~\cite{wikihow} (distant supervision). In our work, we propose to use LLM to create video captions given ASR subtitles, which allows us to create detailed descriptions that are specific for every video and have proper sentence structure.

\subsection{LLMs in Vision-Language Tasks}

In recent years, there has been a remarkable success of LLMs in many language-related tasks~\cite{devlin2018bert,radford2019language,raffel2020exploring}. Latest large language models~\cite{neelakantan2022gpt3.5,taori2023alpaca,vicuna2023,touvron2023llama} have demonstrated excellent zero-shot capabilities on common-sense inference~\cite{chang2023language}. This success has prompted research into integrating common-sense knowledge into vision-language tasks. In this regard, some methods~\cite{sun2019videobert,su2019vl,lu2019vilbert,tan2019lxmert} initialize the language part of vision-language models from pre-trained LLM. Another line of work~\cite{cho2021vlt5,li2023blip,zhao2023learning,liu2024visual} uses LLM as a decoder to enable vision-to-language generation. Works~\cite{lialin2023scalable,zhao2023learning} adapted visually conditioned LLM for visual captioning and created captioning pseudo-labels for large-scale video data. However, these methods require human-annotated datasets to train a captioning model,
while our method does not require any label data and aims to transform free available annotation (ASR subtitles) into textual descriptions.

\section{Method}

\subsection{Problem Statement}

Given a dataset of $N$ untrimmed long-term instructional videos $V_n$ with corresponding noisy ASR
subtitles $S_n$, our goal is to create ``human-written-like'' video captions $C_n$ (with $1 \leq n \leq N$). Note that our task does not assume access to any paired training data $((V_n, S_n), C_n)$. 
The goal is to create the video captions $C_n$ in a \textit{zero-shot} setting given only videos and subtitles $(V_n, S_n)$. 
More formally, for each given video $V_n$, we also have a set of subtitles of spoken text in the video, $S_n = \{s_{n,j}, t^{s}_{n,j}, t^{e}_{n,j}\}_{j \leq |S_n|}$ with their start $t^{s}$ and end timestamps $t^{e}$ recognized by ASR-systems. For each video $V_n$, our goal is to generate dense captions and their timestamps $C_n = \{c_{n,i}, \tau^{s}_{n,i}, \tau^{e}_{n,i} \}_{i \leq |C_n|}$, where each caption ${c_{n,i}}$ describes a segment of the video, that starts at $\tau^{s}_{n,i}$ and ends at $\tau^{e}_{n,i}$. 

The generated captions aim to serve for video-language or video-language-\{other modalities (\eg, audio)\} tasks, providing language supervision in the form of human-style captions rather than scrambled noisy ASR subtitles. That enables the potential of collecting large-scale datasets with long-term videos and their dense textual descriptions for free, without human supervision.

\subsection{Video-Language Retrieval Model}
\label{sec:retr_model}

Before we describe our method for generating the \methodname\ dataset, we will briefly recap the video-language retrieval models (V-L model), as it is one of the main use cases for this dataset. Moreover, we also use a V-L model to improve the temporal alignment in the dataset.

We base our video-language retrieval model (V-L model) on the pre-trained BLIP image-language dual-encoder model~\cite{li2022blip}. We maintain the architecture of the text encoder $f(c) \in \mathbb{R}^d$ but, following CLIP4CLIP~\cite{luo2022clip4clip}, adapt the image encoder $g(I)$ to a video encoder by averaging image embeddings obtained from uniformly sampled frames of the video: $g(V_n) = \sum_{I \in V_n}{g(I)} \in \mathbb{R}^d$.
Dual-encoder models typically learn a cross-modal embedding space~\cite{radford21clip,luo2022clip4clip} via training with the symmetric InfoNCE loss~\cite{oord2018representation}. The training is based on a similarity metric (often cosine distance) between embeddings $\rho_{n,i,m} = \mathrm{sim}\left(f(c_{n,i}), g(V_m)\right)$ scaled by a temperature parameter $\nu$, resulting in the following loss function:
\begin{equation}
\label{eq:loss}
L =-\frac{1}{2|B|}\sum_{(n,i) \in B}\biggl(\log\frac{\exp( \rho_{n,i,n}/\nu )}{\sum\limits_{(m,j) \in B}{\exp(\rho_{n,i,m}/\nu)}}
 + \log\frac{\exp( \rho_{n,i,n}/\nu )}{\sum\limits_{(m,j) \in B}{\exp(\rho_{m,j,n}/\nu)}}\biggr)
\end{equation}
where $B$ is a batch of training sample indices $(n,i)$.

\begin{figure*}[t]
\begin{center}
\includegraphics[width=1\linewidth]{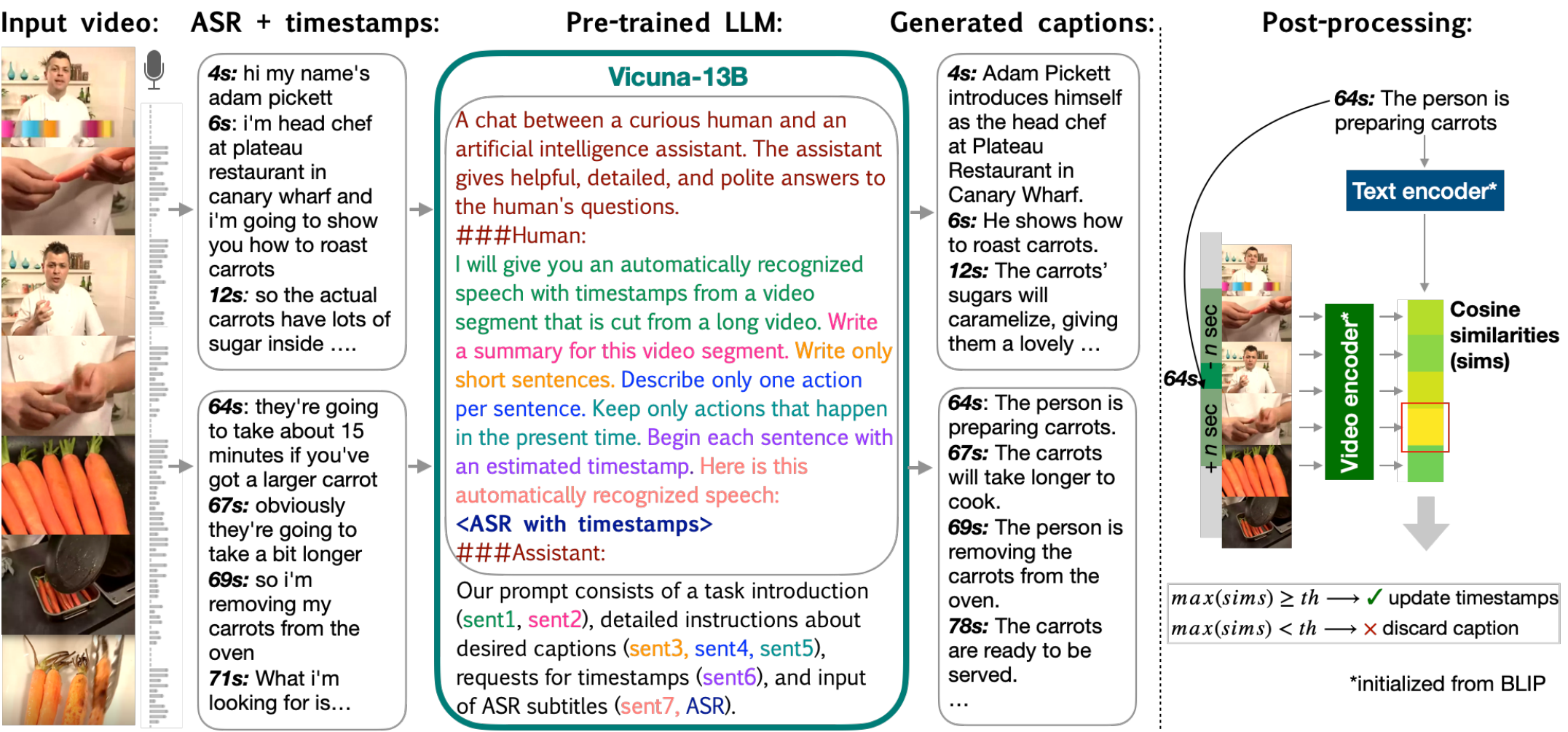}
\end{center}
\caption{\textbf{Schematic visualization of the proposed \methodname\ method}. Obtained from the automatic speech recognition system (ASR), subtitles are divided into blocks that contain longer contextual information. 
A large pre-trained language model is then used to generate plausible video captions based on ASR subtitles, along with timestamps for each caption. These generated captions and timestamps are further post-processed to enhance their alignment to the video and filter out captions with low similarity to the corresponding video by leveraging a pre-trained text-video model.
}
\label{fig:method}
\end{figure*}

\subsection{\methodname~Method}

To generate captions for the instructional videos, we propose to leverage recent large language models
that demonstrate great zero-shot performance in many different tasks formulated with natural language. Namely, we prompt the LLM to read the ASR subtitles of the video and create a plausible video description based on this. 
Since one subtitle only covers a small part of the video and lacks a global context, we propose to aggregate multiple subtitles together with their timestamp information. Then, we task the LLM to create detailed descriptions based on the entire input and estimate timestamps for each generated sentence.

The overview of our approach is shown in~\cref{fig:method}. For each video, we first slice a given sequence of subtitles into blocks that contain long context information about the video. Then, the ASR subtitles of each block are summarised into a video caption using the LLM that we prompt with our task description. 
The LLM also predicts timestamps for each sentence in the video caption, which we further refine in our post-processing step based on similarities of a caption sentence to video clips in the neighboring area of predicted timestamps.

 \myparagraph{LLM Prompting.}
For our language prompt 
(shown in~\cref{fig:method}), we leverage the same ``main'' prompt for the LLM, as in the Vicuna-13B model~\cite{vicuna2023}: ``A chat between a curious human and\dots'' that defines the requirement for LLM to give a helpful answer to our questions. 
Then, we describe our request, what data we need to process, and how it should be processed: ``I will give you an automatically recognized speech\dots''. We found structuring the prompt in the way that the task description given at the beginning of the prompt and the long ASR input $S_n$ at the end is beneficial.  
Then, we give detailed instructions about how to process ASR subtitles.
We found that instructions such as ``Write only short sentences'' or ``Describe only one action per sentence'' are beneficial, as they encourage the creation of concise captions that better match the video content. The instruction ``Keep only actions that happen in the present time'' is intended to filter out unrelated chats, advice, or comments from the captions; we observed that it also resulted in performance enhancements. Lastly, we request the model to predict a timestamp for each generated caption and, finally, input timestamps + ASR subtitles that need to be processed. 
The LLM response follows the start timestamp + caption format given in the prompt and, therefore, can be automatically parsed with a simple script into a set of captions and timestamps $C_n = \{c_{n,i}, \tau^{s}_{n,i}, \tau^{e}_{n,i} \}_{i \leq |C_n|}$, where we assign $\tau^{e}_{n,i} = \tau^{s}_{n,i} + \Delta_{sec}$, where $\Delta_\mathrm{sec}$ is a constant video clip length parameter (number of seconds). 
Please see \cref{sec:ablations} and the supplement for a detailed evaluation of these choices. In the supplement, we also discuss an extension of our HowToCaption method to vision language models that additionally allows grounding captions on visual content. However, we found that this does not improve caption quality.

\myparagraph{Post-processing: Alignment \& Filtering.}
ASR subtitles often temporally misalign with video content~\cite{han2022temporal,miech2019howto100m} (e.g., a speaker describes something only after it was shown in a video). 
Captions derived solely from ASR also face this issue. 
Therefore, inspired by the TAN method~\cite{han2022temporal} that automatically predicts the alignability of subtitles and matching timestamps, we further improve our obtained captions with an alignment \& filtering post-processing step (\cref{fig:method}).
To this end, we utilize the video-language encoder model $(f,g)$. 
Given a generated caption $c_{n,i}$ and its start and end timestamps $(\tau^{s}_{n,i}, \tau^{e}_{n,i})$ that corresponds to a part of the video clip $V_n^{[\tau^{s}_{n,i}, \tau^{e}_{n,i}]}$, we use the V-L model to compute alignment similarity scores $\rho_{n,i}(\delta) = \mathrm{sim}\left(f(c_{n,i}), g(V_n^{[\tau^{s}_{n,i} + \delta, \tau^{e}_{n,i} + \delta]})\right)$ between the caption and video clips with time offsets $\delta \in \mathbb{Z}, |\delta| \leq T$ around predicted timestamps. Then we \textit{align} the caption to the video clip by finding the best offset around the timestamp $\delta^*_{n,i} = \underset{\delta \in \{-T, ..., T\}}{\arg\max} \rho_{n,i}(\delta)$ and \textit{filter} out pairs if $\rho_{n,i}(\delta^*_{n,i}) < \kappa$, where $\kappa$ is a similarity score threshold. 

To further improve the alignment of captions, we perform multiple rounds of alignment \& filtering. In practice, we found that the improvement after two rounds is marginal. For subsequent rounds, we fine-tune 
(c.f.~\cref{sec:retr_model}) 
the V-L model on the aligned \& filtered video-captions pairs $\{(v_i, c_i)\}$, resulting in new alignment scores $\rho'_{n,i}(\delta)$. Since fine-tuning V-L models often leads to forgetting, we employ two modifications in the fine-tuning and second alignment processes. First, during fine-tuning, we add regularization
\begin{equation}
L_\mathrm{align} = \alpha\frac{1}{2|B|}\sum_{(n,i) \in B}(\mathrm{sim}(f(c_{n,i}), f^*(c_{n,i})) + \mathrm{sim}(g(V_n), g^*(V_n))
\end{equation}
where $f^*$ and $g^*$ denote frozen text and video encoders, $\alpha$ is a regularization weight, and $(n,i) \in B$ represents the samples batch $B$. This regularization prevents the model from forgetting~\cite{hou2019learning}. Then, during alignment \& filtering, we use the average of the similarities of the fine-tuned and original model. We show an impact of these modifications in the supplement.

\begin{figure*}[t]

\begin{subfigure}[t]{0.24\linewidth}
        \includegraphics[width=\textwidth]{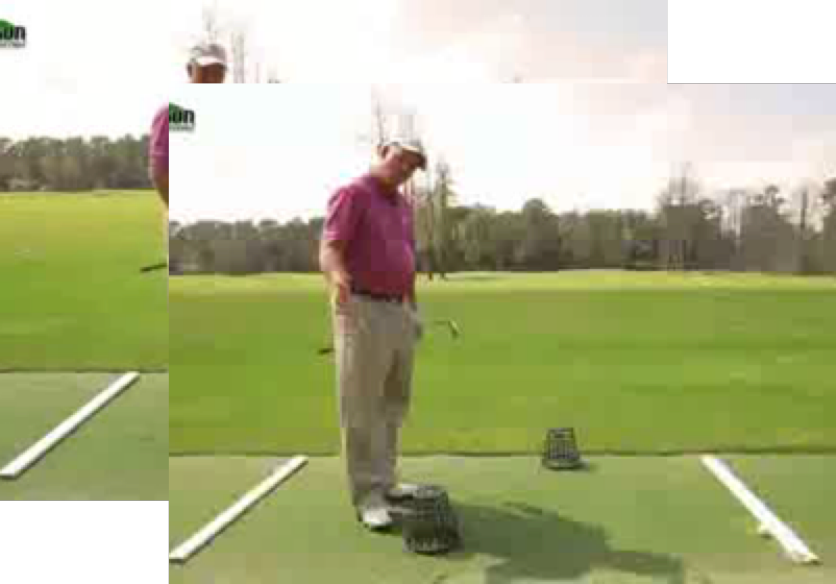}
        \caption*{\scriptsize{\textcolor{gray}{\textbf{ASR}: move them around to help direct the path} \\
        \textbf{Caption}: Matt Swanson gives a tip to use buckets to direct the path of the ball  
        }
        }
\end{subfigure}%
\hfill
\begin{subfigure}[t]{0.24\linewidth}
        \includegraphics[width=\textwidth]{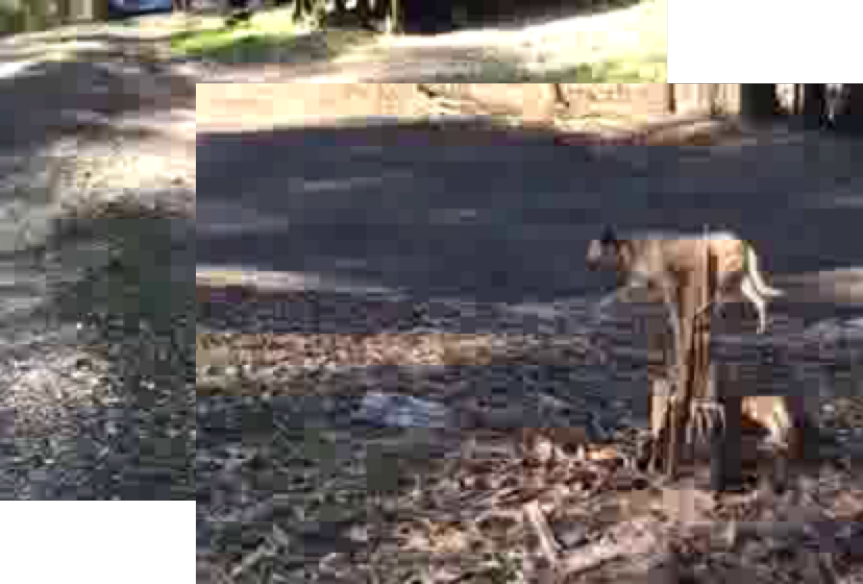}
        \caption*{\scriptsize{\textcolor{gray}{\textbf{ASR}: so this is stage one of hiding the bone burying the bone ...} \\
        \textbf{Caption}: Dog wants to hang out near dirt or other dogs with bones to acquire more bones 
         }
        }
\end{subfigure}%
\hfill
\begin{subfigure}[t]{0.24\linewidth}
        \includegraphics[width=\textwidth]{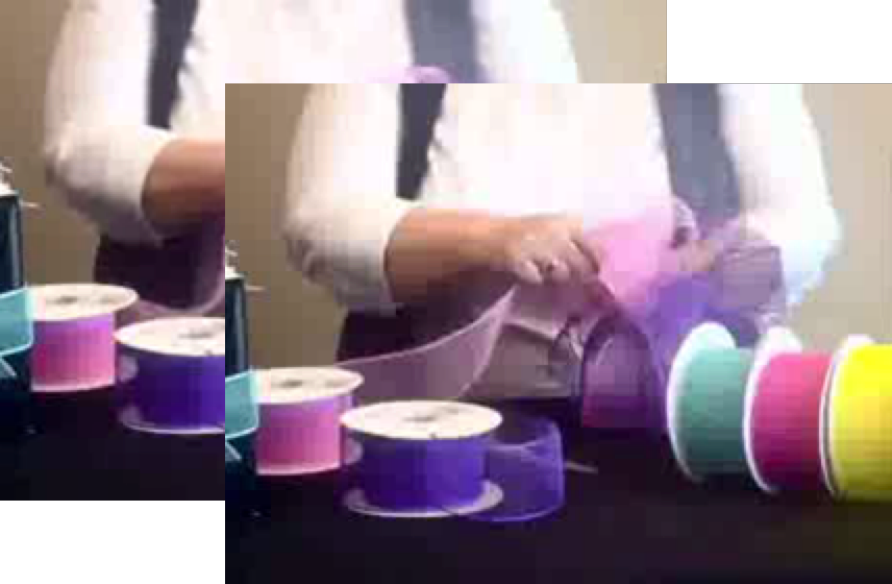}
        \caption*{\scriptsize{\textcolor{gray}{\textbf{ASR}: 
so it's not going to really show }\\
\\
        \textbf{Caption}: Making a bow with two colors
        
}}
\end{subfigure}%
\hfill
\begin{subfigure}[t]{0.23\linewidth}
        \includegraphics[width=\textwidth]{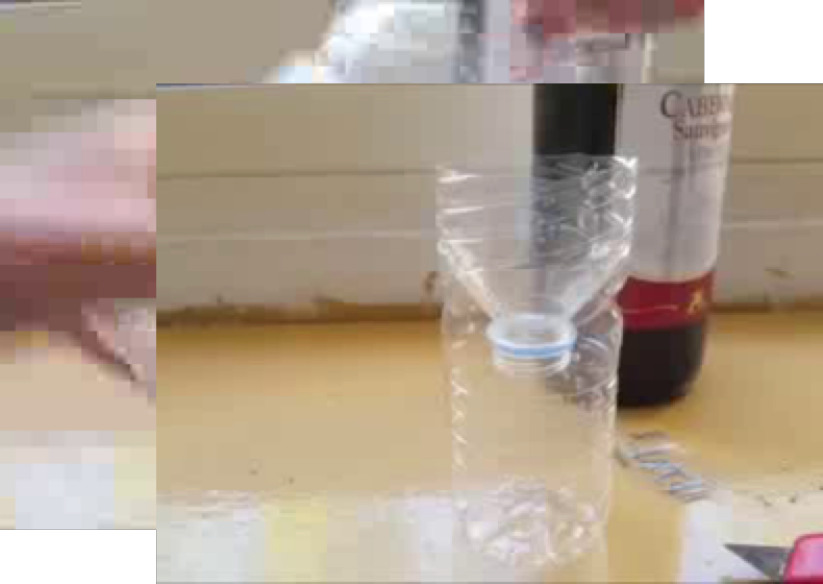}
        \caption*{\scriptsize{\textcolor{gray}{\textbf{ASR}: but this yeah and it just stays or it won't get off it's busy here}\\
        \textbf{Caption}: Make sure the bottle stays together 
        }}
\end{subfigure}%
\caption{\textbf{Examples of video-captions pairs from our \datasetname\ dataset.} ASR subtitles with only noisy supervision for the video are converted from spoken- to written-language-style captions. \fix{Note that some details in the generated captions are taken from a longer context, see the supplement for a full example.} \label{fig:dataset}}
\end{figure*}

\subsection{\datasetname\ Dataset}
\label{sec:dataset}

We apply the proposed \methodname\  approach to 1.2M long-term instructional videos and ASR subtitles of the HowTo100M dataset and obtain the \datasetname\ dataset. By prompting the Vicuna-13B model, we obtain $\sim70$M initial captions.  
\fix{The compute cost for prompting 1.2M videos is $\sim$12k GPU-hours on NVIDIA A40.}
After alignment \& filtering (details in~\cref{sec:impl_details}) we obtain 25M high-quality video-caption pairs. We show examples from our \methodname\ dataset in~\cref{fig:dataset}.  We note that generated captions follow different text styles, \eg, the first and the second examples contain a long description of an object and its actions, the third describes the process, and the last one is instruction. The average length of the captions is 9 words. We provide additional examples, statistics, failure case analyses, and user studies in the supplement.

\section{Experimental Results}

\begin{table*}[t]
    \setlength{\tabcolsep}{2pt}
    \caption{
    \textbf{Ablation of LLM prompts.} We step by step construct a prompt for an LLM that concisely and in detail describes the caption generation task. To emphasize our incremental adjustments, we label the sentences as x$n$ (where $n$ is an index). Each prompt consists of some sentences that were already used in previous prompt versions (e.g., $<$x1$>$, $<$x2$>$) and new sentences introduced in the current prompt (e.g., \textcolor{gray}{x4:} Write only ...). 
    With each prompt, we obtain 2M video-text pairs from 100k HowTo100M videos that we later use for T-V model training (lower-resource setup). Downstream zero-shot text-video retrieval performance is reported. 
    \label{tab:prompts}
    }
    \resizebox{1\linewidth}{!}{
    \footnotesize
    \begin{tabular}{@{}m{7.5cm}|cc|cc|cc|cc|cc@{}}
    	\toprule
                 \multirow{2}{*}{Prompt} & \multicolumn{2}{c}{YouCook2} & \multicolumn{2}{c}{MSR-VTT} &  \multicolumn{2}{c}{MSVD} & \multicolumn{2}{c}{LSMDC} & \multicolumn{2}{c}{Average} \\
                 & R10$\uparrow$ & MR$\downarrow$ & R10$\uparrow$ & MR$\downarrow$  & R10$\uparrow$ & MR$\downarrow$ & R10$\uparrow$ & MR$\downarrow$  & R10$\uparrow$ & MR$\downarrow$ \\
                \midrule 

                \textcolor{gray}{x1:} Here is an automatically recognized speech from a video: $<$ASR with timestamps$>$. \textcolor{gray}{x2:} Write a synopsis for this video. \textcolor{gray}{x3:} Begin each sentence with an estimated timestamp. & 37.5 & 22.5 & 71.0 & \textbf{3} & 80.5 & \textbf{2} & 37.3 & 30 & 56.6 & 14.4 \\
                \midrule
                $<$x1$>$ $<$x2$>$  \textcolor{gray}{x4:} Write only short sentences. $<$x3$>$ & 39.3 & 20.5 & 71.4 & \textbf{3} & 81.0 & \textbf{2} & 36.5 & 32.5 & 57.1 & 14.5  \\ 
                \midrule
                $<$x1$>$ $<$x2$>$  $<$x4$>$ \textcolor{gray}{x5:} Describe only one action per sentence. $<$x3$>$ & 39.8 & 20 & 71.0 & \textbf{3} & 80.9 & \textbf{2} & 37.2 & 30.5 & 57.2 & 13.9 \\ 
                 \midrule
                 $<$x1$>$ $<$x2$>$  $<$x4$>$ $<$x5$>$ \textcolor{gray}{x6:} Keep only actions that happen in the present time. $<$x3$>$ & 39.5 & 19.5 & 71.6 & \textbf{3} & 81.2 & \textbf{2} & \textbf{37.9} & \textbf{29} & 57.6 &\textbf{13.4}   \\
                 \midrule
                 $<$x1$>$ \textcolor{gray}{x2':} Write \textit{a summary} for this video. $<$x4$>$ $<$x5$>$ $<$x6$>$ $<$x3$>$ & 40.4 & \textbf{19} & 71.4 & \textbf{3} & 81.4 & \textbf{2} & 37.1 & 30 & 57.6 & 13.5  \\
                 \midrule
                 \textcolor{gray}{x1':} Here is an automatically recognized speech from \textit{a video segment that is cut from a long video}: $<$ASR with timestamps$>$ \textcolor{gray}{x2'':} Write a summary for \textit{this video segment.} $<$x4$>$ $<$x5$>$ $<$x6$>$ $<$x3$>$ & 40.0 & 20 & \textbf{72.0} & \textbf{3} & 81.1 & \textbf{2} & 37.8 & \textbf{29} & 57.7 & 13.5 \\
                 \midrule
                 \rowcolor{almond} \textit{I will give you} an automatically recognized speech with timestamps from a video segment that is cut from a long video. $<$x2''$>$ $<$x4$>$ $<$x5$>$ $<$x6$>$ $<$x3$>$\textit{ Here is this automatically recognized speech: $<$ASR with timestamps$>$} & \textbf{40.6} & \textbf{19} & \textbf{72.0} & \textbf{3} & \textbf{81.6} & \textbf{2} & 37.7 & 30 & \textbf{58.0} & 13.5 \\ 
         	\arrayrulecolor{black}\bottomrule
    \end{tabular}
    }
\end{table*} 
\begin{table*}[t]
    \centering
    \setlength{\tabcolsep}{1.5pt}
    \caption{
    \textbf{Effect of a longer context.} For the ``no context'' option, we predict captions from individual ASR subtitles. With our ``long context'' option, we input multiple ASR subtitles with timestamps and the model generated captions based on longer context. This ablation is done in lower-resource setup.
    \label{tab:context}
    }
    \resizebox{1\linewidth}{!}{
    \footnotesize
    \begin{tabular}{@{}l|cccc|cccc|cccc|cccc|cccc@{}}
    	\toprule
                 \multirow{2}{*}{Method} & \multicolumn{4}{c}{YouCook2} & \multicolumn{4}{c}{MSR-VTT} &  \multicolumn{4}{c}{MSVD} & \multicolumn{4}{c}{LSMDC} & \multicolumn{4}{c}{Average} \\
                 & R1 & R5 & R10 & MR & R1 & R5 & R10 & MR  & R1 & R5 & R10 & MR  & R1 & R5 & R10 & MR  & R1 & R5 & R10 & MR  \\
                \midrule 
                 No context& 
11.1 & 27.9 & 38.4 & 21 & 37.7 & \textbf{62.4} & \textbf{72.6} & \textbf{3} & 43.3 & 71.7 & 80.2 & \textbf{2} & 16.5 & 30.4 & \textbf{38.4} & \textbf{30} & 27.1 & 48.1 & 57.4 & 14 \\
                 \rowcolor{almond} Long context & \textbf{12.1} & \textbf{30.0} & \textbf{40.6} & \textbf{19} & \textbf{37.9} & 61.6 & 72 & \textbf{3} & \textbf{43.9} & \textbf{72.7} & \textbf{81.6} & \textbf{2} & \textbf{16.8} & \textbf{31.4} & 37.7 & \textbf{30} & \textbf{27.7} & \textbf{48.9} & \textbf{58.0} & \textbf{13.5} \\
         	\arrayrulecolor{black}\bottomrule
    \end{tabular}
    }

\end{table*} 
\begin{table*}[t]
    \centering
    \setlength{\tabcolsep}{2pt}
    \caption{
    \textbf{Effect of alignment \& filtering.}  With each post-processing variant, we obtain 25M video-text pairs that we later use for T-V model training. Downstream zero-shot text-video retrieval performance is reported. }
    \label{tab:post-processing}
    \resizebox{1\linewidth}{!}{
    \footnotesize
    \begin{tabular}{@{}l|cc|cc|cc|cc|cc@{}}
    	\toprule
                 \multirow{2}{*}{Caption Post-processing} & \multicolumn{2}{c}{YouCook2} & \multicolumn{2}{c}{MSR-VTT} &  \multicolumn{2}{c}{MSVD} & \multicolumn{2}{c}{LSMDC} & \multicolumn{2}{c}{Average} \\
                  & R10$\uparrow$ & MR$\downarrow$ & R10$\uparrow$ & MR$\downarrow$  & R10$\uparrow$ & MR$\downarrow$ & R10$\uparrow$ & MR$\downarrow$  & R10$\uparrow$ & MR$\downarrow$ \\
                \midrule 
                 \textcolor{gray}{Lower bound: original ASR as supervision} & \textcolor{gray}{39.3} & \textcolor{gray}{20} & \textcolor{gray}{61.7} & \textcolor{gray}{5} & \textcolor{gray}{77.1} & \textcolor{gray}{2} & \textcolor{gray}{31.5} & \textcolor{gray}{56} & \textcolor{gray}{52.4} & \textcolor{gray}{20.8} \\
                 
                \midrule 
                 No post-processing & 
40.2 & 18 & 65.9 & 4 & 79.8 & \textbf{2}  & 34.4 & 40 & 55.1 & 16.0 \\

                 Filtering (using BLIP) & 42.5 & 16 &  71.2 & \textbf{3}   & 81.7 & \textbf{2}  & 37.4 & 30 & 58.2 & 12.8 \\
                 
                 Alignment \& filtering (using BLIP) & 42.4 & 17 & 71.7 & \textbf{3}  & \textbf{82.2} & \textbf{2} & 38.5 & 29.5 & 58.7 & 12.9 \\
                 
                 \rowcolor{almond} Alignment \& filtering (with ours) & \textbf{44.1} & \textbf{15} & \textbf{73.3} & \textbf{3} & 82.1 & \textbf{2}  & \textbf{38.6} & \textbf{29} & \textbf{59.5} & \textbf{12.3} \\
         	\arrayrulecolor{black}\bottomrule
    \end{tabular}
    }
\end{table*}

To evaluate the proposed \datasetname\ dataset for large-scale pre-training of vision-language models, we train a T-V model as described in~\cref{sec:retr_model} on the \datasetname\ dataset and assess its zero-shot video-text retrieval performance on four widely recognized and diverse video-text benchmarks: YouCook2~\cite{zhou2018towards}, MSR-VTT~\cite{xu2016msr}, MSVD~\cite{chen2011collecting}, and LSMDC~\cite{rohrbach2015long}. While the YouCook2 dataset consists of instructional cooking videos and might be considered as an in-domain benchmark for the \datasetname\ dataset, the other datasets encompass a broader range of topics and video types, including non-instructional YouTube videos and movies. To evaluate the properties of \datasetname\ dataset in comparison with other large-scale pre-training datasets,
we also train our T-V model on the HowTo100M~\cite{miech2019howto100m}, HowTo100M with step labels~\cite{lin2022learning}, HTM-AA~\cite{han2022temporal}, VideoCC3M~\cite{nagrani2022learning}, and WebVid2M~\cite{Bain21} datasets and compare zero-shot text-video retrieval performance. 

\subsection{Datasets and Metrics}

\textbf{Pre-training Datasets.}  \textbf{HowTo100M} is a dataset of 1.2M instructional videos with ASR subtitles. 
We consider three versions of annotations of this dataset: \textit{Sentencified HowTo100M}~\cite{han2022temporal}, with pre-processed ASR subtitles by structuring them into full sentences;
\textit{HowTo100M with Distant Supervision}~\cite{lin2022learning}, where ASR subtitles were linked to WikiHow~\cite{wikihow} step descriptions via distant supervision; 
and \textit{HTM-AA}~\cite{han2022temporal}, an auto-aligned (AA) version of HowTo100M, where subtitle timestamps were adjusted to improve alignment to videos and non-alignable subtitles were discarded. 
\textbf{WebVid2M}~\cite{Bain21} is a large open-domain dataset of 2.5M of short videos scrapped from the internet with their alt-text. \textbf{VideoCC3M}~\cite{nagrani2022learning} is a dataset of 10M video-text pairs collected by transferring captions from image-text CC3M dataset~\cite{changpinyo2021conceptual} to videos with similar visual content. 

\myparagraph{Datasets for Downstream Tasks.} \textbf{YouCook2}~\cite{zhou2018towards} is a dataset of instructional cooking videos, where each video clip is annotated with a recipe step. We used 3.5k test set for evaluation.
\textbf{MSR-VTT}~\cite{xu2016msr} contains 10k YouTube videos on various topics and human descriptions. Following prior work~\cite{Bain21,nagrani2022learning,tang2021clip4caption}, we use the 1k test set for evaluation. \textbf{MSVD}~\cite{chen2011collecting} is a dataset of video snippets with their textual summary. The evaluation set consists of 670 videos with 40 captions corresponding to each video. We follow standard practice~\cite{Bain21,luo2022clip4clip} and count each caption-video pair towards the metrics. \textbf{LSMDC}~\cite{rohrbach2015long} is a collection of video clips from movies with human-written descriptions. The test set consists of 1k video-caption pairs. 

\myparagraph{Metrics.} To evaluate zero-shot text-video retrieval, we use standard Recall@$K$  metrics where $K \in{1, 5, 10}$ (R1, R5, R10) and Median Rank (MR).

\subsection{Implementation Details}
\label{sec:impl_details}
As an LLM, we utilize Vicuna-13B-v0~\cite{vicuna2023}.
In the supplement, we additionally experiment with the MiniGPT-4 model~\cite{zhu2023minigpt4} to generate captions from subtitles grounded on visual content. To create the \datasetname\ dataset,  we leverage subtitles with timestamps released by~\cite{han2022temporal} (Sentencified HowTo100M), where officially released subtitles~\cite{miech2019howto100m} for HowTo100M videos were post-processed by structuring them into full sentences.
For our T-V model (described in~\cref{sec:retr_model}), we use ViT-B/16 visual encoder and $\textrm{BERT}_{\textrm{base}}$ textual encoder that are initialized with $\textrm{BLIP}_{\textrm{CapFilt-L}}$ pre-trained weights.
We uniformly sample 4 frames from a video clip during training and 12 frames during evaluation.
For  \methodname\ method, we use $T=10$ seconds offset for alignment and adaptive threshold $\kappa$ to leave 25M most similar pairs after filtering. Following~\cite{chen2021multimodal} that found that 8-sec clips are optimal for training on HowTo100M, we set $\Delta_\mathrm{sec}=8$. More implementation details are in the supplement.

\subsection{Ablation Studies}\label{sec:ablations}

\textbf{Prompt Engineering.} First, we assess the quality of the obtained dataset with respect to different LLM prompts.
Since prompting LLM with subtitles from 1.2M videos is resource-intensive, we perform the prompt engineering ablations in a lower-resource setup, where we use a 100k subset of HowTo100M ($\sim$~10\% of all videos) to create dense captions with the LLM and use the threshold $\kappa$ to obtain the 2M most confident video-caption pairs.  
Here, we train the T-V model for 150k iterations and then evaluate zero-shot on downstream tasks.  In \cref{tab:prompts}, we begin with a basic prompt for the LLM, gradually refining it to generate captions more suitable for vision-language tasks. It is essential to recognize that the impact of various prompts on performance can vary across datasets, as certain prompts may yield captions better aligned with specific downstream tasks. Notably, incorporating key phrases such as ``Write only short sentences'' or ``Describe only one action per sentence'' leads to performance improvements on 3 out of 4 datasets. Additionally, the use of the phrase ``Keep only actions that happen in the present time'' 
also results in performance enhancements. Furthermore, structuring the task description at the beginning and presenting the data to be processed at the end (the final modification) also boosts performance. We provide more ablations on prompt engineering in the supplement.

We also examine the impact of leveraging a longer context for caption prediction. In~\cref{tab:context}, we compare caption generation with ``no context'', where captions are predicted from individual ASR subtitles. 
With our ``long context'' option, we input multiple ASR subtitles with their timestamps, and the model predicts both captions and timestamps based on longer context. We found that using a longer context is beneficial, resulting in an average improvement of $0.6$ p.p. in R10, and particularly advantageous for YouCook2 and MSVD.

\begin{table*}[t]
    \centering
    \setlength{\tabcolsep}{1.5pt}
    
    \caption{
    \textbf{Zero-shot text-to-video retrieval performance of models trained on different video-text datasets.} For each dataset, we train our T-V model and report downstream 
    performance.
    \label{tab:training_datasets}
    }
    \resizebox{1\linewidth}{!}{
    \footnotesize
    \begin{tabular}{@{}l|cccc|cccc|cccc|cccc@{}}
    	\toprule
                 \multirow{2}{*}{Video-Text Training Data} & \multicolumn{4}{c}{YouCook2} & \multicolumn{4}{c}{MSR-VTT} &  \multicolumn{4}{c}{MSVD} & \multicolumn{4}{c}{LSMDC}  \\
                  & R1 & R5 & R10 & MR & R1 & R5 & R10 & MR & R1 & R5 & R10 & MR & R1 & R5 & R10 & MR  \\
                \midrule 
                  \rowfont{\color{gray}} - (zero-shot, with BLIP initialization)  & \rowfont{\color{gray}} 6.1 & \rowfont{\color{gray}}16.2 & \rowfont{\color{gray}}23.6 & \rowfont{\color{gray}}69 & \rowfont{\color{gray}}34.3 & \rowfont{\color{gray}}59.8 & \rowfont{\color{gray}}70.6 & \rowfont{\color{gray}}3 & \rowfont{\color{gray}}38.5 & \rowfont{\color{gray}}65.0 & \rowfont{\color{gray}}74.0 & \rowfont{\color{gray}}2 & \rowfont{\color{gray}}14.7 & \rowfont{\color{gray}}29.5 & \rowfont{\color{gray}}36.5 & \rowfont{\color{gray}}31 \\
                 HowTo100M with ASRs~\cite{han2022temporal} & 12.2 & 29.1 & 39.3 & 20 & 30.8 & 52.6 & 61.7 & 5 & 39.2 & 68.3 & 77.1 & \textbf{2} & 12.9 & 24.7 & 31.5 & 56  \\
                 HowTo100M with distant supervision~\cite{lin2022learning} & 8.3 & 21.5 & 30.3 & 34 & 28.6 & 54.0 & 66.3 & 5 & 38.5 & 68.6 & 79.4 & \textbf{2} & 12.1 & 24.7 & 32.4 & 42.5 \\
                 HTM-AA (auto-aligned ASRs)~\cite{han2022temporal} &  \textbf{13.4} & 32.2 & 43.5 & \textbf{15} & 29.8 & 54.1 & 64.3 & 4 & 38.7 & 68.6 & 78.7 & \textbf{2} & 11.9 & 23.9 & 30.5 & 46  \\
                 
                 \rowcolor{almond} \datasetname\ (ours)  & \textbf{13.4} & \textbf{33.1} & \textbf{44.1} & \textbf{15} & 37.6 & \textbf{62.0} & \textbf{73.3} & \textbf{3} & \textbf{44.5} & 73.3 & \textbf{82.1} & \textbf{2} & 17.3 & \textbf{31.7} & 38.6 & 29 \\
                 \midrule
                 VideoCC3M~\cite{nagrani2022learning} & 5.3 & 15.1 & 21.7 & 84 & 33.9 & 57.9 & 67.1 & 4 & 39.6 & 66.7 & 76.8 & \textbf{2} & 14.8 & 29.4 & 35.8 & 33 \\
                 WebVid2M~\cite{Bain21} & 7.3 & 20.7 & 29.0 & 46 & \textbf{38.5} & 61.7 & 71.9 & \textbf{3} & 44.5 & \textbf{73.4} & 82.1 & \textbf{2} & \textbf{17.8} & 31.2 & \textbf{39.8} & \textbf{25}  \\
         	\arrayrulecolor{black}\bottomrule
    \end{tabular}
    }
    
\end{table*}

\begin{table*}[t]
    \centering
    \setlength{\tabcolsep}{1.5pt}
    
    \caption{
    \textbf{Comparison in zero-shot text-to-video retrieval with \textit{dual-encoder} baseline methods.} 
      $^{\S}$For BLIP, the performance of dual-encoder architecture is reported. $^\ddagger$6 datasets = CC3M~\cite{changpinyo2021conceptual}+CC12M~\cite{changpinyo2021conceptual}+COCO~\cite{lin2014microsoft}+VG~\cite{krishna2017visual} +SBU~\cite{ordonez2011im2text} +LAION~\cite{schuhmann2021laion}. ``V.'' = Vision, ``I-T'' = Image-Text, ``V-T'' = Video-Text.
    \label{tab:sota}
    }
    \resizebox{1\linewidth}{!}{
    \footnotesize
    \begin{tabular}{@{}lccc|cccc|cccc|cccc|cccc@{}}
    	\toprule
                 \multirow{2}{*}{Method} & \multirow{2}{*}{V. Encoder} & \multirow{2}{*}{I-T Data} & \multirow{2}{*}{V-T Data} & \multicolumn{4}{c}{YouCook2} & \multicolumn{4}{c}{MSR-VTT} &  \multicolumn{4}{c}{MSVD} & \multicolumn{4}{c}{LSMDC} \\
                 & & & & R1 & R5 & R10 & MR & R1 & R5 & R10 & MR   & R1 & R5 & R10 & MR  & R1 & R5 & R10 & MR \\
                \midrule 
                 Nagrani et al.~\cite{nagrani2022learning} & ViT-B & - & VideoCC3M & - & - & - & - & 18.9 & 37.5 & 47.1 & - & - & - & -& - & - & - & - \\
                 
                 Frozen-in-Time~\cite{Bain21} & ViT-B/16 & CC+COCO  & WebVid2M & - & - & - & - & 24.7 & 46.9 & 57.2 & 7 \\
                 
                 CLIP-straight~\cite{portillo2021straightforward} & ViT-B/32 & WIT & - & - & - & - & - & 31.2 & 53.7 & 64.2 & 4 & 37.0 & 64.1 & 73.8 & 2 & 11.3 & 22.7 & 29.2 & 56.5 \\
                 
                 CLIP4CLIP~\cite{luo2022clip4clip} & ViT-B/32 & WIT & HowTo100M &  - & - & - & - & 32.0 & 57.0 & 66.9 & 4 & 38.5 & 66.9 & 76.8 & 2 & 15.1 & 28.5 & 36.4 & \textbf{28} \\
                 
                 VideoCoCa~\cite{yan2022video} & ViT-B/18& JFT-3B & VideoCC3M &  \textbf{16.5} & - & - & - & 31.2 & - & - & - & - & - & - & - & - & - & - & -  \\ 
                 
                 BLIP$^{\S}$~\cite{li2022blip} & ViT-B/16 & 6 datasets$^\ddagger$ & - & 6.1 & 16.2 & 23.6 & 69 & 34.3 & 59.8 & 70.6 & \textbf{3} & 38.5 & 65.0 & 74.0 & 2 & 14.7 & 29.5 & 36.5 & 30.5 \\

                 BLIP$^{\S}$+HowTo100M & ViT-B/16 & 6 datasets$^\ddagger$ & HowTo100M & 12.2 & 29.1 & 39.3 & 20 & 30.8 & 52.6 & 61.7 & 5 & 39.2 & 68.3 & 77.1 & 2 & 12.9 & 24.7 & 31.5 & 56 \\
                 
                 \rowcolor{almond} \textbf{Ours} & ViT-B/16 & 6 datasets$^\ddagger$ & HowToCaption & 13.4 &\textbf{ 33.1} & \textbf{44.1} & \textbf{15} & \textbf{37.6} & \textbf{62} & \textbf{73.3} & \textbf{3} & \textbf{44.5} & \textbf{73.3} & \textbf{82.1} & \textbf{2}& \textbf{17.3} & \textbf{31.7} & \textbf{38.6} & 29 \\

         	\arrayrulecolor{black}\bottomrule

    \end{tabular}
    }
    
\end{table*}

\begin{table}[t]
    \centering
    \footnotesize
    \setlength{\tabcolsep}{3pt}
    \caption{
    \textbf{Zero-shot text-to-video+audio retrieval.} $^\ast$Text-video only models.}
    \label{tab:t2va}
    \resizebox{0.75\linewidth}{!}{
    \begin{tabular}{@{}lc|cccc|cccc@{}}
    	\toprule
                 \multirow{2}{*}{Method} & \multirow{2}{*}{Vision Encoder} & \multicolumn{4}{c}{YouCook2} & \multicolumn{4}{c}{MSR-VTT} \\
                  &&  R1$\uparrow$ & R5$\uparrow$ & R10$\uparrow$ & MR$\downarrow$ & R1$\uparrow$ & R5$\uparrow$ & R10$\uparrow$ & MR$\downarrow$  \\
                \midrule 
                 MIL-NCE$^\ast$~\cite{miech2020end}  & S3D & 15.1 & 38.0 & 51.2 & 10 & 9.9 & 24.0 & 32.4 & 29.5\\
                 TAN$^\ast$~\cite{han2022temporal} & S3D & 20.1 & 45.5 & 59.5 & 7.0 & - & - & - & - \\
                 MMT~\cite{gabeur2020multi} & Transformer & - & - & - & - & - & 14.4 & - & 66 \\
                 AVLNet~\cite{rouditchenko2020avlnet} & ResNet-152+ResNeXt101 & 19.9 &  36.1 & 44.3 & 16 &  8.3 & 19.2 & 27.4  & 47 \\
                 MCN~\cite{chen2021multimodal} & ResNet-152+ResNeXt101 & 18.1 & 35.5 & 45.2 & - & 10.5 & 25.2 & 33.8 & - \\
                 EAO~\cite{shvetsova2022everything} & S3D & 24.6 & 48.3 & 60.4 & 6 & 9.3 & 22.9 & 31.2 & 35 \\
                 \rowcolor{almond} Ours  & S3D & \textbf{25.5} & \textbf{51.1} & \textbf{63.6} & \textbf{5} & \textbf{13.2} & \textbf{30.3} & \textbf{41.5} & \textbf{17} \\
         	\arrayrulecolor{black}\bottomrule
    \end{tabular}
    }
\end{table} 

\begin{table*}[t]
    \centering
    \setlength{\tabcolsep}{1.5pt}
    \caption{
    \textbf{Video captioning results.} 
We report BLEU@4 (B@4), METEOR (M), ROUGE-L (R), and CIDEr (C).   We  \textcolor{gray}{gray out}
methods that use a stronger vision backbone and significantly more pre-training data for fair comparison.  $^\star$4 datasets = CC3M+COCO+VG+SBU, $^\dagger$5 datasets = ...+CC12M, $^\ddagger$6 datasets = ...+LAION~\cite{schuhmann2021laion}. $^{\S}$Our fine-tuning.  ``V.'' = Vision, ``I-T'' = Image-Text, ``V-T'' = Video-Text.
    \label{tab:captioning}
    }
    \resizebox{1\linewidth}{!}{
    \footnotesize
    \begin{tabular}{@{}lccc|cccc|cccc|cccc@{}}
    	\toprule
                 \multirow{2}{*}{Method} & \multirow{2}{*}{V. Encoder} & \multirow{2}{*}{I-T Data} & \multirow{2}{*}{V-T Data} & \multicolumn{4}{c}{YouCook2} & \multicolumn{4}{c}{MSR-VTT} &  \multicolumn{4}{c}{MSVD} \\
                 & & & & B@4 & M & R & C & B@4 & M & R & C & B@4 & M & R & C \\
                \midrule 
                SwinBERT~\cite{lin2022swinbert}& VidSwin-B & - & - &\textbf{9.0}&15.6&37.3&109 & 41.9&29.9&62.1&53.8&58.2&41.3&77.5&120.6\\
                CLIP4Caption~\cite{tang2021clip4caption} & ViT-B	& -	& - & - & - & - & - & 46.1&30.7 & 63.7 &57.7& - & - & - & -  \\
                GIT-B~\cite{wang2022git} & ViT-B & 4 datasets$^\star$ & - & 5.8 & 12.2 & 31.5 & 80.3 & 46.6 & 29.6 & 63.2 & 57.8 & 69.3 & 44.5 & 81.4 & 142.6  \\
                MV-GPT~\cite{seo2022end} & ViViT-B & - & HowTo100M& - & - & - & -   & 48.9&\textbf{38.6}&64&60.0& - & - & - & - \\
                LAVENDER & VidSwin-B & 5 datasets$^\dagger$ & WebVid2M+12M & - & - & - & -  & - & - & - & 60.1 & - & - & - & 150.7   \\
                HiTeA~\cite{ye2023hitea} & MViT-B &  5 datasets$^\dagger$ & WebVid2M & - & - & - & - &  - & - & - & 65.1 & - & - & - & 146.9    \\
                mPlug-2~\cite{xu2023mplug} & ViT-B &  5 datasets$^\dagger$ & WebVid2M& - & - & - & - & \textbf{52.2} & 32.1&\textbf{66.9}&\textbf{72.4}&69.3&45.1&81.9&148.2   \\
                BLIP~\cite{li2022blip} & ViT-B	& 6 datasets$^\ddagger$ & - & 7.9 & 15.0 & 36.0 & 104.8 & 49.2 & 32.2 & 66.1 & 65.5 & 68.0 & 45.3 & 82.8 & 148.6  \\
                BLIP + HowTo100M$^{\S}$ & ViT-B & 6 datasets$^\ddagger$ & HowTo100M & 8.6 & \textbf{15.9} & 37.1 & 112.9 & 49.4 & 32.0 & 66.2 & 65.9 & 69.3 & 46.2 & 83.0 & 151.4  \\
                \rowcolor{almond} Ours & ViT-B	& 6 datasets$^\ddagger$ & HowToCaption & 8.8 & \textbf{15.9} & \textbf{37.3} & \textbf{116.4} & 49.8 & 32.2 & 66.3 & 65.3 & \textbf{70.4} & \textbf{46.4} & \textbf{83.2} & \textbf{154.2}  \\
                \midrule
                \textcolor{gray}{Vid2Seq~\cite{yang2023vid2seq}} & \textcolor{gray}{ViT-L} & \textcolor{gray}{-} & \textcolor{gray}{YT-Temporal-1B} & \textcolor{gray}{-} & \textcolor{gray}{-} & \textcolor{gray}{-} & \textcolor{gray}{-} & \textcolor{gray}{-} & \textcolor{gray}{30.8} & \textcolor{gray}{-} & \textcolor{gray}{64.6} & \textcolor{gray}{-} & \textcolor{gray}{45.3} & \textcolor{gray}{-} & \textcolor{gray}{146.2}  \\
                
                \textcolor{gray}{VideoCoCa~\cite{yan2022video}} & \textcolor{gray}{ViT-G} & \textcolor{gray}{JFT-3B} & \textcolor{gray}{VideoCC3M} & \textcolor{gray}{-} & \textcolor{gray}{-} & \textcolor{gray}{-} & \textcolor{gray}{-} & \textcolor{gray}{53.8} & \textcolor{gray}{-} & \textcolor{gray}{68.0} & \textcolor{gray}{73.2} & \textcolor{gray}{-} & \textcolor{gray}{-} & \textcolor{gray}{-} & \textcolor{gray}{-} \\
                
                \textcolor{gray}{GIT-2~\cite{wang2022git}} & \textcolor{gray}{DaViT-4.8B} & \textcolor{gray}{12.9B pairs} & \textcolor{gray}{-} & \textcolor{gray}{9.4} & \textcolor{gray}{15.6} & \textcolor{gray}{37.5} & \textcolor{gray}{131.2} & \textcolor{gray}{54.8} & \textcolor{gray}{33.1} & \textcolor{gray}{68.2} & \textcolor{gray}{75.9} & \textcolor{gray}{82.2} & \textcolor{gray}{52.3} & \textcolor{gray}{88.7} & \textcolor{gray}{185.4}  \\

         	\arrayrulecolor{black}\bottomrule

    \end{tabular}
    }
\end{table*}

\myparagraph{Alignment \& Filtering.} Further, we assess the impact of the proposed alignment \& filtering procedure on the quality of captions of the acquired dataset in~\cref{tab:post-processing}. We examine the performance of the T-V model when trained on differently post-processed versions of the dataset. Remarkably, we discover that the obtained video-caption pairs, even without any post-processing, significantly outperform the original ASR-based supervision. Subsequently, by employing the alignment and filtering procedure to leave only 25M pairs based on video-caption similarities derived from BLIP pre-trained weights, we achieve a notable performance enhancement of 3.6 p.p. in R10. Furthermore, alignment \& filtering with our proposed fine-tuning without forgetting yields an additional 0.8 p.p. boost in R10 performance. More ablations can be found in the supplement.

\subsection{Comparison with State-of-the-art}

\textbf{Comparision with Other Web Datasets.} In \cref{tab:training_datasets}, we assess the pre-training effectiveness of our proposed \datasetname\ dataset compared to other web video-language datasets. Specifically, we evaluate different textual annotations of HowTo100M videos: sentencified ASR subtitles~\cite{han2022temporal},  task steps from distant supervision~\cite{lin2022learning}, and auto-aligned ASR subtitles~\cite{han2022temporal}. Additionally, we conduct evaluations on WebVid2M~\cite{Bain21} and VideoCC3M~\cite{nagrani2022learning} datasets. Our findings indicate that the model pre-trained on our \datasetname\ dataset significantly outperforms models pre-trained on other versions of HowTo100M annotations, with an average improvement of 5.2 p.p. in R10. This improvement is most pronounced for the MSR-VTT, MSVD, and LSMDC datasets, which feature full-sentence captions. Interestingly, for the YouCook2 dataset with captions in the form of step descriptions like ``cut tomato'', HTM-AA already exhibits a high baseline performance, but our \datasetname\ dataset still provides a performance boost.
We also observe that the VideoCC3M dataset does not improve the initial BLIP performance on any datasets except for the MSVD. We attribute it to the fact that the VideoCC3M  dataset adopts captions from the CC3M dataset~\cite{changpinyo2021conceptual} and transfers them to videos, potentially not introducing significantly new knowledge for the BLIP-initialised model since BLIP was pre-trained on multiple datasets, including CC3M. On the other hand, WebVid2M demonstrated performance improvements across all datasets, but our \datasetname\ dataset notably outperforms WebVid2M on YouCook2 and MSR-VTT, only underperforming on LSMDC.

\myparagraph{Comparison with State-of-the-art in Zero-shot Text-Video Retrieval.} In \cref{tab:sota}, we also conduct a comparison with zero-shot \textit{dual-encoder} retrieval baselines. It is important to acknowledge that comparing state-of-the-art methods can be challenging due to variations in backbone capacity, training objectives, and other factors. Therefore, we focus on a comparison with dual-encoder models in the zero-shot settings. Additional comparisons with methods that use \textit{re-ranking} can be found in the supplement. Nevertheless, it is worth highlighting that our approach consistently outperforms the baseline methods in zero-shot text-video retrieval across all datasets.

\myparagraph{Zero-shot Text-to-Video+Audio Retrieval.} It is known that instructional video datasets, \eg, HowTo100M, suffer from a high correlation of audio modality to a textual description, therefore hindering building a text-video+audio retrieval system where the video is extended with audio~\cite{shvetsova2022everything,nagrani2022learning}. The usage of ASR narrations as supervisory textual description leads retrieval models to primarily perform speech recognition on the audio, hindering true language-audio connections~\cite{shvetsova2022everything}.
Therefore, training text-video+audio systems on such datasets usually requires additional regularization, such as shifting audio timestamps or assigning lower weights to the audio loss~\cite{shvetsova2022everything}.  Our \datasetname\ dataset resolves this issue by providing richer textual descriptions, allowing us to train a text-video+audio retrieval system without regularization.  To evaluate this, we train a multimodal Everything-At-Once (EAO)~\cite{shvetsova2022everything} model that learns to fuse any combinations of text, video, and audio modalities on our proposed \datasetname\ dataset without any additional regularization and evaluate zero-shot text-video+audio retrieval performance. \cref{tab:t2va} shows the proposed model significantly outperforms all baselines and the directly comparable EAO model.

\myparagraph{Video Captioning.}
We further validate our dataset's potential for video-language pre-training through video captioning. We fine-tune the BLIP~\cite{li2022blip} model with all architecture blocks, including the image-grounded text decoder, on the HowToCaption dataset with all three BLIP losses: image-text contrastive loss, image-text matching loss, and language modeling loss. We adapt the image encoder to the video encoder similarly to our retrieval model~(\cref{sec:retr_model}) and use similar hyperparameters (see the supplement). 
Subsequently, we fine-tune the model with language modeling loss on the training set of either YouCook2, MSR-VTT, or MSVD datasets and evaluate video captioning performance.
\cref{tab:captioning}~show that our model fine-tuned on the HowToCaption dataset significantly outperforms directly comparable BLIP model and other comparable baselines on YouCook2 and MSVD and keeps competitive performance on MSR-VTT.

\section{Conclusion}

In this work, we propose a novel approach, HowToCaption, that transforms freely available ASR subtitles of instructional videos into high-quality captions, enabling the collection of large-scale, high-quality video-language datasets without manual annotation efforts. 
With this approach, we curate a new large-scale HowToCaption dataset featuring 
human-style video captions derived from ASR subtitles of the HowTo100M dataset. 
We demonstrate that our HowToCaption dataset serves as an excellent source for training video-language representation and video captioning models across four text-video retrieval and three video captioning benchmarks. Furthermore, given the separation of textual descriptions from the audio modality in our dataset, it demonstrates efficacy as a valuable resource for text-to-video-audio tasks. This work demonstrates the potential of LLMs for creating annotation-free, large-scale video-language datasets.

\section*{Acknowledgements}
Nina Shvetsova is supported in part by the German Federal Ministry of Education and Research (BMBF) project STCL - 01IS22067.

\end{sloppypar}
\bibliographystyle{splncs04}
\bibliography{egbib}

\clearpage

\addcontentsline{toc}{section}{Appendix} 
\appendix
\noindent{\Large\bf Supplementary Material}\\[1em]
\numberwithin{table}{section}
\numberwithin{figure}{section}

\fix{In the appendix, we provide additional experimental results, implementation and dataset details, and qualitative examples. Additionally, we discuss limitations and responsibility to human subjects. 
Specifically, we first discuss the experimental results of using MiniGPT-4~\cite{zhu2023minigpt4} to generate visually grounded captions in Sec.~\ref{sec:grounding}.
Then, we provide additional experimental results in Sec.~\ref{sec:additionalexps}, including prompt engineering experiments in Sec.~\ref{sec:additionalpropmpts}, ablations of our filtering \& alignment method in Sec.~\ref{sec:ablationfiltalign}, and robustness analysis in Sec.~\ref{sec:robustness}. Then, we provide additional implementation details in Sec.~\ref{sec:additional_implementation_details}, statistics of the HowToCaption dataset in Sec.~\ref{sec:dataset_details}, and qualitative examples in Sec.~\ref{sec:dataset_examples}. Finally, we discuss our method limitations in Sec.~\ref{sec:limitations} and responsibility to human subjects in Sec.~\ref{sec:responsibility}.}

\section{Grounding Captions to Video Content with MiniGPT-4}\label{sec:grounding}

Our generated captions with Vicuna-13B are based solely on ASR subtitles. To additionally ground the produced captions on visual content, we experiment with the recent MiniGPT-4 model~\cite{zhu2023minigpt4}. The MiniGPT-4 consists of the frozen Vicuna-13B model and a visual encoder with a Q-Former~\cite{li2022blip} that projects visual features from an image into tokens in a language model embedding space that are later treated as word tokens in the Vicuna-13B model. 
To ground generated captions on the visual modality, we create a grid image from 4 uniformly sampled frames from a video clip and slightly adapt the prompt to encourage the LLM to utilize the provided image for generating captions (\cref{tab:final_prompt}). We apply our approach to obtain visually grounded captions with the MiniGPT-4 model and obtain \textit{\datasetname-grounded}. For this dataset, we follow exactly the same hyperparameters that we use for \datasetname. 
In~\cref{tab:LLM}, we evaluate the downstream retrieval performance of the T-V model trained on \datasetname-grounded. The dataset shows mixed results compared to the \datasetname; while it is beneficial for the MSR-VTT and the MSVD dataset, performance on the YouCook2 dataset drops. To facilitate further analysis,  we will release both caption sets: the ASR-based only \datasetname, produced by the Vicuna-13B, and \datasetname-grounded, produced by the MiniGPT-4. 

\begin{table*}[t]
    \setlength{\tabcolsep}{3pt}
    \caption{\small{
    \textbf{Prompts for the Vicuna-13B and MiniGPT-4 models.} Difference is highlighted with bold.}
    \label{tab:final_prompt}
    }
    \vspace{-0.25cm}
    \resizebox{1\linewidth}{!}{
    \footnotesize
    \begin{tabular}{@{}m{7cm}|m{7cm}@{}}
    	\toprule
                 Vicuna-13B & MiniGPT-4 \\
                \midrule 
                 I will give you an automatically recognized speech with timestamps from a video segment that is cut from a long video. Write a summary for this video segment. Write only short sentences. Describe only one action per sentence. Keep only actions that happen in the present time. Begin each sentence with an estimated timestamp. Here is this automatically recognized speech: $<$ASR with timestamps$>$ & I will give you an automatically recognized speech with timestamps and an image with four frames from a video segment that is cut from a long video. Write a summary for this video segment \textbf{based on both: video frames and speech.} Write only short sentences. Describe only one action per sentence. Keep only actions that happen in the present time. Begin each sentence with an estimated timestamp. \textbf{Here is the image with four frames: $<$Img$>$$<$grid-image here$>$$<$/Img$>$}. Here is the automatically recognized speech: $<$ASR with timestamps$>$ \\
         	\arrayrulecolor{black}\bottomrule
    \end{tabular}
    }
    \vspace{-0.25cm}
\end{table*} 
\begin{table*}[t]
    \centering
    \footnotesize
    \setlength{\tabcolsep}{1.5pt}
    \caption{\small{
    \textbf{Comparison of \datasetname\ and \datasetname-grounded datasets obtained with Vicuna-13b and MiniGPT-4 large language models, respectively.} For each dataset, we train a T-V model and report downstream zero-shot text-video retrieval performance. }
    \label{tab:LLM}
    }
    \vspace{-0.25cm}
    \resizebox{1\linewidth}{!}{
    
    \begin{tabular}{@{}l|cccc|cccc|cccc|cccc@{}}
    	\toprule
                 \multirow{2}{*}{Dataset} & \multicolumn{4}{c}{YouCook2} & \multicolumn{4}{c}{MSR-VTT} &  \multicolumn{4}{c}{MSVD} & \multicolumn{4}{c}{LSMDC}  \\
                  & R1 & R5 & R10 & MR & R1 & R5 & R10 & MR   & R1 & R5 & R10 & MR & R1 & R5 & R10 & MR \\
                \midrule 
                 \datasetname\ (Vicuna-13B) & 13.4 & 33.1 & 44.1 & 15 & 37.6 & 62.0 & 73.3 & 3 & 44.5 & 73.3 & 82.1 & 2 & 17.3 & 31.7 & 38.6 & 29 \\
                 \datasetname-grounded (MiniGPT-4) & 12.4 & 29.8 & 39.9 & 20.5 & 38.3 & 62.5 & 73.2 & 2 & 46.2 & 73.9 & 82.5 & 2 & 16.8 & 31.0 & 38.7 & 27 \\
         	\arrayrulecolor{black}\bottomrule
    \end{tabular}
    }
    \vspace{-0.25cm}
\end{table*} 

\section{Additional Experimental Evaluation}\label{sec:additionalexps}

\subsection{Dataset Quality}

\begin{table*}[t]
    \centering
    \setlength{\tabcolsep}{1.5pt}
    \caption{\small{
    \textbf{ Caption quality evaluation in the HowTo100M and HowToCaption datasets using the HIREST dataset as the ground truth.} We use a subset of 423 HIREST video clip-caption pairs that temporally overlap with video clip-caption pairs in both the HowToCaption and HowTo100M datasets with > 0.5 IoU. We report standard text generation metrics. }
    \label{tab:text_quality}
    }
    \resizebox{0.70\columnwidth}{!}{
		 \begin{tabular}{@{}l|cccc@{}}
    	\toprule
                 & BLEU@4 & METEOR & ROUGE-L & CIDEr \\
                 \midrule 
                 HowTo100M with ASRs & 0.4 & \textbf{9.8} & 11.4 & 3.1 \\
                 \rowcolor{almond} HowToCaption (ours) & \textbf{1.0} & {9.3} &\textbf{ 13.6} & \textbf{29.6} \\
         	\bottomrule
    \end{tabular}
		}
\end{table*}

In this section, we provide an additional quantitative evaluation of the quality of captions in our HowToCaption dataset. 

\myparagraph{Evaluation Using the HIREST Dataset.} First, we evaluate the caption quality using the HIREST dataset~\cite{zala2023hierarchical} as the ground truth. The HIREST dataset contains step captions for a small subset of video clips from the HowTo100M dataset. Since the video clip boundaries of the HIREST dataset, our HowToCaption dataset, and the HowTo100M dataset differ, we use a subset of 423 HIREST (video clip, caption) pairs that have corresponding (video clip', caption') and (video clip'', caption'') pairs in the HowToCaption and HowTo100M datasets respectively, such as the temporal boundaries of (video clip and video clip') overlaps with > 0.5 IoU, as well as (video clip and video clip'') overlaps with > 0.5 IoU. Using HIREST captions as ground truth, we report the quality of captions in the HowTo100M and HowToCaption datasets using standart metrics such as BLEU@4 and METEOR, as shown in ~\cref{tab:text_quality}. Despite the common disadvantages of classical metrics, which focus more on wording rather than semantics, we observe consistent improvement in the captions of our HowToCaption dataset over the ASR subtitles.

\begin{table*}[]
    \setlength{\tabcolsep}{2pt}
    \caption{\small{
    \textbf{Additional experiments with LLM prompts.}  We report modifications that we have done in comparison to our default prompt, which is \colorbox{almond}{highlighted}. With each prompt, we obtain 2M video-text pairs from 100k HowTo100M videos that we later use for T-V model training (low-recourse setup). Downstream zero-shot text-video retrieval performance is reported. }
    \label{tab:prompts_add}
    }
    \vspace{-0.25cm}
    \resizebox{1\linewidth}{!}{
    \footnotesize
    \begin{tabular}{@{}m{7.5cm}|cc|cc|cc|cc|cc@{}}
    	\toprule
                 \multirow{2}{*}{Prompt} & \multicolumn{2}{c}{YouCook2} & \multicolumn{2}{c}{MSR-VTT} &  \multicolumn{2}{c}{MSVD} & \multicolumn{2}{c}{LSMDC} & \multicolumn{2}{c}{\textbf{Average}} \\
                 & R10$\uparrow$ & MR$\downarrow$ & R10$\uparrow$ & MR$\downarrow$  & R10$\uparrow$ & MR$\downarrow$ & R10$\uparrow$ & MR$\downarrow$  & R10$\uparrow$ & MR$\downarrow$ \\
                \midrule 
                 \rowcolor{almond}I will give you an automatically recognized speech with timestamps from a video segment that is cut from a long video. Write a summary for this video segment. Write only short sentences. Describe only one action per sentence. Keep only actions that happen in the present time. Begin each sentence with an estimated timestamp. Here is this automatically recognized speech: $<$ASR with timestamps in the format ``n''s: ``ASR''$>$  \textbf{(ours)} & 40.6 & 19 & 72.0 & 3 & 81.6 & 2 & 37.7 & 30 & 58.0 & 13.5 \\ 
                 \midrule
                 Modification:  Write a summary for this video segment. $\longrightarrow$ Write a \textbf{likely} summary for this video segment.  & 40.8 & 18.5 & 71.4 & 3 & 81.5 & 2 & 37.7 & 30 & 57.9 & 13.4 \\
                 \midrule
                 Modification:  Write a summary for this video segment. $	\longrightarrow$ Write a \textbf{creative} summary for this video segment. & 40.0 & 19 & 71.6 & 3 & 81.2 & 2 & 37.8 & 27 & 57.7 & 12.8 \\ 
                 \midrule
                 Modification: $<$ASR with timestamps in the format ``n''s: ``ASR''$>$ $\longrightarrow$ $<$ASR with timestamps in the format \textbf{``minutes'':``seconds'': ``ASR subtitle''}$>$  & 40.8 & 18.5 & 71.5 & 3 & 81.2 & 2 & 37.2 & 29 & 57.7 & 13.1\\ 
         	\arrayrulecolor{black}\bottomrule
    \end{tabular}
    }
    \vspace{-0.25cm}
\end{table*} 

\myparagraph{User Study.} To quantitatively evaluate the ``human-written''-like quality of our captions, we performed a user study with 60 random captions: 20 from HowToCaption, 20 ASR subtitles, and 20 captions from the downstream datasets, namely YouCook2, MSRVTT, MSVD, and LSMDC (5 captions per dataset), which served as a control group. We asked users to identify \textit{human-written video captions}. Using majority voting from 30 participants, 90\% of the control group, 50\% of HowToCaption, and only 15\% of ASR captions were chosen as ``human-written'' captions. This demonstrates that our captions are significantly more ``human-written'' than the ASR subtitles.

\subsection{Additional Comparisons with SOTA in Text-Video Retrieval}

\begin{table*}[t]
    \centering
    \setlength{\tabcolsep}{1.5pt}
    \caption{
    \textbf{Comparison of zero-shot text-to-video retrieval methods with re-ranking.} We include methods that perform a re-ranking of the top-N retrieved candidates using a visual-text matching head. $^\star$$\textrm{BLIP}_{\textrm{CapFilt-L}}$ (initialization of our model). $^\dagger$Text-video model only (without audio and subtitles).
    \label{tab:sota_reranking}
    }
    \resizebox{1\linewidth}{!}{
    \footnotesize
    \begin{tabular}{@{}lc|cccc|cccc|cccc|cccc@{}}
    	\toprule
                 \multirow{2}{*}{Method} & \multirow{2}{*}{V. Encoder}  & \multicolumn{4}{c}{YouCook2} & \multicolumn{4}{c}{MSR-VTT} &  \multicolumn{4}{c}{MSVD} & \multicolumn{4}{c}{LSMDC} \\
                 & & R1 & R5 & R10 & MR & R1 & R5 & R10 & MR   & R1 & R5 & R10 & MR  & R1 & R5 & R10 & MR \\
                \midrule 
BLIP$^\star$~\cite{li2022blip} & ViT-B & 10.7 & 24.1 & 32.3 & 44.5 & 41.4 & 63.3 & 72.8 & 2 & 45.6 & 71.3 & 79.7 & 2 & 20.5 & 36.5 & 44.2 & 18.5\\
\rowcolor{almond} BLIP$^\star$, HowToCaption-finetuned & ViT-B & 18.15 & 38.68 & 50.4 & 10 & 44.3 & 66.6 & 76.6 & 2 & 49 & 76.2 & 84.2 & 2 & 19.6 & 34.3 & 42.9 & 18\\
BLIP, COCO-finetuned~\cite{li2022blip} & ViT-B & - & - & - & - & 43.3 & 65.6 & 74.7 & 2 & 48.7 & 75.8 & 83.1 & 2 & 22.3 & 38.4 & 44.1 & 18\\
Unmasked Teacher-17M~\cite{li2023unmasked} & ViT-L & - & - & - & - & 42.6 & 64.4 & 73.1 & - & 49.9 & 77.7 & 85.3 & - & 25.2 & 43 & 50.5 & -\\
mPLUG, COCO-finetuned~\cite{li2022mplug} & ViT-L & - & - & - & - & 44.3 & 66.4 & 75.4 & - & - & - & - & - & - & - & - & -\\
mPLUG-2~\cite{xu2023mplug} & ViT-L & - & - & - & - & 47.1 & 69.7 & 79 & - & - & - & - & - & 24.1 & 43.8 & 52 & -\\
VAST~\cite{chen2024vast} & ViT-G & - & - & - & - & 49.3 & 68.3 & 73.9 & 2 & - & - & - & - & - & - & - & -\\
VAST$^\dagger$~\cite{chen2024vast} & ViT-G & 15.7 & 35.2 & 45.6 & 14 & 48.5 & 71.2 & 79.8 & 2 & 50.6 & 76.2 & 84.1 & 1 & 23.2 & 40.9 & 48.9 & 12\\
\rowcolor{almond}VAST$^\dagger$, HowToCaption-finetuned & ViT-G & \textbf{19.7} & \textbf{43.6} & \textbf{53.9} & \textbf{8} & \textbf{50} & \textbf{73.2} & \textbf{81.4} & \textbf{1} & \textbf{54.8} & \textbf{80.9} & \textbf{87.2} & \textbf{1} & \textbf{27.7} & \textbf{46.5} & \textbf{54.6} & \textbf{7}\\
         	\arrayrulecolor{black}\bottomrule

    \end{tabular}
    }
    
\end{table*}

In addition to comparing dual encoder-only models in zero-shot text-to-video retrieval, in~\cref{tab:sota_reranking}, we compare methods that use a visual-text matching head for \textit{re-ranking} the best candidates~\cite{li2022mplug,chen2024vast}. Specifically, we evaluate our model with all BLIP architecture blocks used in video captioning evaluation (as described in Sec. 4.4 of the main paper and in~\cref{sec:additional_implementation_details}) by re-ranking 128 best candidates from dual encoder model predictions using a visual-text matching head. Moreover, we fine-tune the state-of-the-art text-video VAST model~\cite{chen2024vast} for 9k interactions on the HowToCaption dataset with a batch size of 64, a learning rate of 5e-06, using 4 frames per video clip, and following other training/evaluation parameters of VAST. We observe that fine-tuning on the HowToCaption dataset boosts the performance of both models. Furthermore, the text-video VAST model fine-tuned on the HowToCaption dataset outperforms all other models.

\subsection{Additional Results in Prompt Engineering}\label{sec:additionalpropmpts}

In~\cref{tab:prompts_add}, we provide an additional evaluation of language prompts. First, we experiment with phrases such as ``write a \textit{likely} summary$\ldots$'' and ``write a \textit{creative} summary$\ldots$''. While the keyword ``likely'' almost does not change downstream performance, the keyword ``creative'' is not beneficial for 3 out of 4 datasets. We also experiment with another timestamp format in the LLM prompt. Namely, instead of using \textit{``n''s} (such as 0s, 65s), we use  \textit{``minutes'':``seconds''} format (such as 00:00, 01:05). We found that simple timestamp format \textit{``n''s} results in a higher performance. 

\subsection{Ablations of Filtering \& Alignment Post-processing} \label{sec:ablationfiltalign}

\begin{table*}[t]
    \centering
\caption{\small{
    \textbf{Ablation of our alignment \& filtering method.} With each post-processing variant, we obtain a dataset that we later use for T-V model training. Downstream zero-shot text-video retrieval performance is reported. Options used to obtain the main results are \colorbox{almond}{highlighted}.}
    \label{tab:alignment_filtering_abl}
}
\vspace{-0.5cm}
\begin{subtable}[t]{1\linewidth}
   \centering
    \caption{\small{
    \textbf{Components of our filtering \& alignment method.}} \label{tab:filtering_abl}}
    \vspace{-0.25cm}
    \resizebox{1\linewidth}{!}{
    \footnotesize
    \begin{tabular}{@{}m{7cm}|cc|cc|cc|cc|cc@{}}
    	\toprule
                 \multirow{2}{*}{Caption post-processing} & \multicolumn{2}{c}{YouCook2} & \multicolumn{2}{c}{MSR-VTT} &  \multicolumn{2}{c}{MSVD} & \multicolumn{2}{c}{LSMDC} & \multicolumn{2}{c}{\textbf{Average}} \\
                  & R10$\uparrow$ & MR$\downarrow$ & R10$\uparrow$ & MR$\downarrow$  & R10$\uparrow$ & MR$\downarrow$ & R10$\uparrow$ & MR$\downarrow$  & R10$\uparrow$ & MR$\downarrow$ \\ 
                 \midrule 
                 Alignment \& filtering (using the BLIP) & 42.4 & 17 & 71.7 & 3 & 82.2 & 2 & 38.5 & 29.5 & 58.7 & 12.9 \\
                 \midrule 
                  Alignment \& filtering after second round & 42.4 & 17 & 69.4 & 3 & 81.2 & 2 & 38.1 & 33 & 57.8 & 13.8 \\
                  + regularization $L_{align}$ & 
44.3 & 15 & 71.9 & 3 & 81.9 & 2 & 39.0 & 28 & 59.3 & 12.0  \\
                  + averaging similarities of the finetuned and original model  & 
43.7 & 15 & 72.8 & 3 & 82.0 & 2 & 39.6 & 27 & 59.5 & 11.8 \\
                  \rowcolor{almond} + regularization $L_{align}$ + averaging similarities of the finetuned and original model \textbf{(ours)} & 44.1 & 15 & 73.3 & 3 & 82.1 & 2 & 38.6 & 29 & 59.5 & 12.3 \\
         	\arrayrulecolor{black}\bottomrule
    \end{tabular}
    }
\end{subtable}

\begin{subtable}[t]{1\linewidth}
\caption{\small{
    \textbf{Rounds of alignment \& filtering with finetuning of the T-V model after each round.}} \label{tab:filtering_more_rounds}}
    \vspace{-0.25cm}
    \resizebox{1\linewidth}{!}{
    \footnotesize
    \begin{tabular}{@{}m{7cm}|cc|cc|cc|cc|cc@{}}
    	\toprule
                 \multirow{2}{*}{Caption post-processing} & \multicolumn{2}{c}{YouCook2} & \multicolumn{2}{c}{MSR-VTT} &  \multicolumn{2}{c}{MSVD} & \multicolumn{2}{c}{LSMDC} & \multicolumn{2}{c}{\textbf{Average}} \\
                  & R10$\uparrow$ & MR$\downarrow$ & R10$\uparrow$ & MR$\downarrow$  & R10$\uparrow$ & MR$\downarrow$ & R10$\uparrow$ & MR$\downarrow$  & R10$\uparrow$ & MR$\downarrow$ \\ 
                 \midrule 
                 Alignment \& filtering (using the BLIP) = 1 round & 42.4 & 17 & 71.7 & 3 & 82.2 & 2 & 38.5 & 29.5 & 58.7 & 12.9  \\
                 \rowcolor{almond}Alignment \& filtering after 2'nd round \textbf{(ours)} & 44.1 & 15 & 73.3 & 3 & 82.1 & 2 & 38.6 & 29 & 59.5 & 12.3 \\
                 Alignment \& filtering after 4'th round & 
44.5 & 15 & 72.2 & 3 & 81.8 & 2 & 38.6 & 29 & 59.3 & 12.3 \\
         	\arrayrulecolor{black}\bottomrule
    \end{tabular}
    }
\end{subtable} 

\begin{subtable}[t]{1\linewidth}
\centering
    \setlength{\tabcolsep}{1pt}
    \caption{\small{
    \textbf{Range of time offsets during alignment.}} \label{tab:offset}}
    \vspace{-0.25cm}
    \resizebox{1\linewidth}{!}{
    \footnotesize
    \begin{tabular}{@{}l|cccc|cccc|cccc|cccc|cccc@{}}
    	\toprule
                 Alignment time offsets & \multicolumn{4}{c}{YouCook2} & \multicolumn{4}{c}{MSR-VTT} &  \multicolumn{4}{c}{MSVD} & \multicolumn{4}{c}{LSMDC} & \multicolumn{4}{c}{\textbf{Average}} \\
                  $\delta \in \mathbb{Z}, |\delta| \leq T$ & R1 & R5 & R10 & MR & R1 & R5 & R10 & MR & R1 & R5 & R10 & MR & R1 & R5 & R10 & MR & R1 & R5 & R10 & MR \\
                \midrule 
                  $T = 0$ seconds (no align.) &  12.9 & 32.1 & 43.5 & 15 & 38 & 61.5 & 71.9 & 3 & 44.1 & 72.9 & 81.9 & 2 & 16.8 & 31.0 & 37.8 & 30.5 & 28.0 & 49.4 & 58.8 & 12.6  \\
                 \rowcolor{almond} $T = 10$ seconds & 13.4 & 33.1 & 44.1 & 15 & 37.6 & 62.0 & 73.3 & 3 & 44.5 & 73.3 & 82.1 & 2 & 17.3 & 31.7 & 38.6 & 29 & 28.2 & 50.0 & 59.5 & 12.3 \\
                $T = 20$ seconds & 13.2 & 32.0 & 42.9 & 16 & 37.9 & 62.4 & 72.8 & 3 & 44.8 & 73.4 & 82.2 & 2 & 16.7 & 31.9 & 39.0 & 28 & 28.2 & 49.9 & 59.2 & 12.3 \\
         	\arrayrulecolor{black}\bottomrule
    \end{tabular}}
\end{subtable}

\begin{subtable}[t]{1\linewidth}
\centering
    \setlength{\tabcolsep}{1pt}
    \caption{\small{
    \textbf{Filtering threshold $\kappa$.}} \label{tab:pair_number}}
    \vspace{-0.25cm}
    \resizebox{1\linewidth}{!}{
    \footnotesize
    \begin{tabular}{@{}l|cccc|cccc|cccc|cccc|cccc@{}}
    	\toprule
                 Number of pairs after  & \multicolumn{4}{c}{YouCook2} & \multicolumn{4}{c}{MSR-VTT} &  \multicolumn{4}{c}{MSVD} & \multicolumn{4}{c}{LSMDC} & \multicolumn{4}{c}{\textbf{Average}} \\
                   filtering (varying $\kappa$)  & R1 & R5 & R10 & MR & R1 & R5 & R10 & MR & R1 & R5 & R10 & MR & R1 & R5 & R10 & MR & R1 & R5 & R10 & MR \\
                \midrule 
                  15M pairs & 13.5 & 32.2 & 43.0 & 16.5 & 37.6 & 62.2 & 72.6 & 3 & 45.1 & 73.4 & 82.2 & 2 & 16.7 & 32.3 & 38.5 & 28.5 & 28.2 & 50.0 & 59.1 & 12.5 \\
                 \rowcolor{almond} 25M pairs & 13.4 & 33.1 & 44.1 & 15 & 37.6 & 62.0 & 73.3 & 3 & 44.5 & 73.3 & 82.1 & 2 & 17.3 & 31.7 & 38.6 & 29 & 28.2 & 50.0 & 59.5 & 12.3 \\
                 40M pairs & 12.9 & 32.7 & 43.7 & 15 & 37.8 & 60.5 & 71.3 & 3 & 44.1 & 73.3 & 81.8 & 2 & 16.7 & 31.9 & 37.8 & 29.5 & 27.9 & 49.6 & 58.7 & 12.4 \\
         	\arrayrulecolor{black}\bottomrule
    \end{tabular}
    }    
\end{subtable} 

\end{table*}

We present ablations of our alignment \& filtering post-processing in \cref{tab:alignment_filtering_abl}. In \cref{tab:filtering_abl}, we ablate two modifications of the fine-tuning and alignment processes for the second round of filtering \& alignment. We observe that the dataset obtained after the second round of filtering \& alignment without these modifications shows lower performance than the dataset obtained with the first round (using the BLIP model). We attribute this to forgetting during fine-tuning. However, we note that both proposed modifications boost performance, as well as their combination. 

In~\cref{tab:filtering_more_rounds}, we also analyze if more than two rounds of filtering \& alignment lead to a better quality dataset. We employ 20k iterations of fine-tuning of the T-V model on the obtained dataset after each filtering \& alignment round. We do not observe any performance boost with more filtering \& alignment rounds. 

We further ablate the range of alignment time offsets $T$ and filtering threshold $\kappa$ used in our alignment \& filtering method in~\cref{tab:offset} and \cref{tab:pair_number} respectively. We found that alignment with up to $T=10$ seconds offsets and filtering with the threshold that selects 25M most similar video-text pairs result in the highest performance.

\subsection{Robustness Analysis} \label{sec:robustness}

\fix{We further analyze the robustness of our HowToCaption method to noisy input. First, we examine how the method performs if the video channel is corrupted, specifically containing only empty black frames. We replace all 1.2M videos with videos of the same length but containing only black frames while keeping the original ASR subtitles of the videos. Consequently, the generated captions from subtitles remain the same, but the videos are corrupted. After our alignment \& filtering procedure, we find that only $\sim$1.8\% of generated captions are matched. Therefore, our method filters out almost all captions for corrupted videos.  Furthermore, we test the case of data where ASR subtitles do not match the videos. For each video among the 1.2M videos, we randomly assign ASR subtitles from another video of the dataset. In this case, generated captions from subtitles are the same, but they do not correspond to the videos. We find that after applying the HowToCaption method to such input data, only $\sim$2.4\% of captions are matched after alignment \& filtering. Therefore, almost all noisy and irrelevant captions are filtered, indicating our method's robustness for noisy data.}

\section{Additional Implementation Details} \label{sec:additional_implementation_details}

For our T-V model, we follow BLIP's~\cite{li2022blip} dual encoder architecture with a ViT-B/16 visual encoder and a $\textrm{BERT}_{\textrm{base}}$ textual encoder, which are initialized with $\textrm{BLIP}_{\textrm{CapFilt-L}}$ pre-trained weights.
Following BLIP~\cite{li2022blip}, we also use an extension of the loss
(Equation 1 in the paper)
with soft labels produced by a momentum encoder and a memory bank that keeps additional text and video embeddings from the previous iterations. We train the model for 300k iterations using AdamW~\cite{loshchilov2017decoupled} with a batch size of 128, a learning rate of 1e-6, and a weight decay of 0.05. We use a memory bank of 2048 and smooth labels with a parameter of 0.6. Training augmentation is cropping with a scale [0.5, 1].  For model fine-tuning in the alignment \& filtering step, we use 20k training iterations and regularization parameter $\alpha=0.1$.

\myparagraph{Video Captioning Details.} \fix{For our video captioning experiments, we fine-tune the BLIP~\cite{li2022blip} model with all architecture blocks, including the image encoder, the text encoder, the image-grounded text encoder, and the image-grounded text decoder. Following~\cite{li2022blip}, we use all three BLIP losses: image-text contrastive loss, image-text matching loss, and language modeling loss for fine-tuning. If not stated otherwise, we use the same hyperparameters as in our main text-video retrieval experiments, including the ViT-B/16 visual encoder and the $\textrm{BERT}_{\textrm{base}}$ textual encoder initialized with the $\textrm{BLIP}_{\textrm{CapFilt-L}}$ weights, as well as the same learning rate, batch size, etc.} 

\fix{We fine-tune the model for 200k iterations on the HowToCaption dataset. Then, we fine-tune the image captioning model: the image encoder and the image-grounded text decoder (following~\cite{li2022blip}) on the corresponding training set of the YouCook2, MSRVTT, or MSVD datasets. Note that the MSRVTT dataset has different splits for retrieval and captioning evaluation. Following standard 
practice~\cite{lin2022swinbert,tang2021clip4caption,seo2022end}, 
we use 6.5k videos for training and 3.5k for testing of video captioning. For fine-tuning on downstream datasets, we use a batch size of 16, a learning rate of 1e-5, and a learning rate scheduler with cosine decay. We sample 16 frames for training and evaluation. We fine-tune for 10 epochs on the YouCook2, 30 epochs on the MSRVTT, and 40 epochs on the MSVD. For captioning performance evaluation, we set a number of beams = 1 and a maximum length = 20, as in~\cite{lin2022swinbert}.}

\myparagraph{Text-to-Video+Audio Retrieval Details.} \fix{For text-to-video+audio experiments, we train a multimodal Everything-At-Once (EAO) model~\cite{shvetsova2022everything} with frozen S3D features~\cite{miech2020end} on our HowToCaption dataset. Since our dataset addresses the issues of high correlation between audio and text, we exclude additional regularizations that aim to prevent the model from learning shortcuts, i.e., simplifying the task to speech recognition from audio while ignoring the video.
by simply performing speech recognition in audio and ignoring video.
Specifically, we omit shifting audio timestamps with respect to video clip boundaries and assigning lower weights to the audio loss. All hyperparameters are kept the same as in EAO\cite{shvetsova2022everything}. }

\section{HowToCaption Dataset Statistics} \label{sec:dataset_details}

\begin{table*}[t]
    \centering
    \setlength{\tabcolsep}{1.5pt}
    \caption{\small{
    \textbf{ Language statistics.} $|V|$ is the vocabulary size. \#word/caption is the number of words per caption. \#word/video is the number of words per all captions in a video. \%diverse verb is the percentage of diverse verbs. All numbers are obtained from 5000 randomly sampled videos. }
    \label{tab:statistics}
    }
    \vspace{-0.25cm}
    \resizebox{1\linewidth}{!}{
    \scriptsize
    \begin{tabular}{@{}m{1.9cm}|ccc|c|ccc|ccc@{}}
    	\toprule
                 \multirow{3}{*}{Dataset} & \multicolumn{3}{c|}{Standard statistics} & \multicolumn{1}{c|}{Diversity$\uparrow$} & \multicolumn{3}{c|}{$n$-grams diversity$\uparrow$} & \multicolumn{3}{c}{Verb $n$-grams diversity$\uparrow$} \\
                  & \multirow{2}{*}{$|V|$}  & \#word  & \#word  & \multirow{2}{*}{\%diverse verbs} & \multirow{2}{*}{1-gram} & \multirow{2}{*}{2-gram}& \multirow{2}{*}{3-gram}& \multirow{2}{*}{1-gram} & \multirow{2}{*}{2-gram}& \multirow{2}{*}{3-gram}\\ 
                  & & /caption  & /video  &  &  & &&  & & \\ 
                 \midrule 
                ASR subtitles    & 45905  &  10.96  & 909.34           & 77.09            & 1.01     & 17.57   & 50.24    & \textbf{1.13}          & 19.88        & 61.64         \\ \midrule
                HowToCaption & 36204   &   9.03   & 581.27             & \textbf{82.55}            & \textbf{1.25}     & \textbf{21.36}   & \textbf{53.95}    & 1.01          & \textbf{25.9}         & \textbf{76.65}         \\ 
         	\arrayrulecolor{black}\bottomrule
    \end{tabular}
    }
    
\end{table*}

In this section, we present the statistics of our \datasetname\ dataset. Our goal is to demonstrate the scale and diversity of the captions in the proposed dataset. 
 
\begin{figure*}[t]

\begin{subfigure}[t]{0.49\linewidth}
        \includegraphics[width=\textwidth]{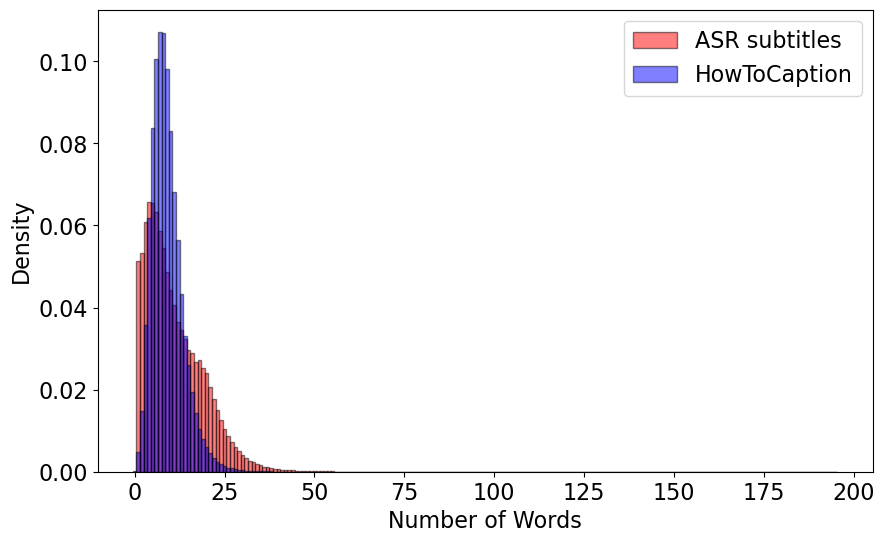}
        \caption*{\footnotesize{Histogram of caption lengths. }}
\end{subfigure}%
\hfill
\begin{subfigure}[t]{0.49\linewidth}
        \includegraphics[width=\textwidth]{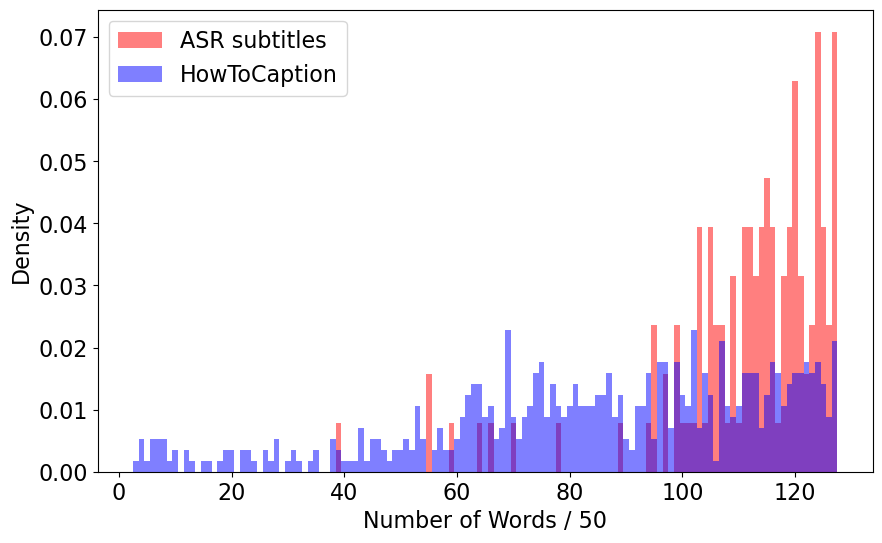}
        \caption*{\footnotesize{Histogram of total length of all captions for a video.  The scale of the x-axis is divided by 50. }}
\end{subfigure}%
\vspace{-0.25cm}
\caption{ \small{\textbf{Caption length statistics of our \datasetname\ dataset.} We randomly sample 5000 videos to plot the distributions. \label{fig:statistics}}}
\vspace{-0.25cm}

\end{figure*}

\myparagraph{Caption Length.}
To better understand the scale of our dataset, we compute caption length statistics. We analyze captions both at the video clip level and at the video level (when combining captions from all clips belonging to the same video). We randomly sample 5000 videos from \datasetname\ and use a spaCy tokenizer~\cite{honnibal2020spacy} to count words. The resulting histograms of caption length are shown in \cref{fig:statistics} and statistics in \cref{tab:statistics}. On a sentence level, our dataset has shorter captions on average (9.03 words) compared to the original ASR subtitles (10.97 words). Our captions also have a smaller standard deviation (4.36 vs. 7.91), indicating a more consistent length distribution.

\myparagraph{Language Diversity.} We also compare the language diversity of the ASR subtitles and our captions. We measure language diversity from two perspectives:  1) diversity based on the presence of distinct words or verbs and 2) diversity of word/verb n-grams across the captions, providing insights into the varied combinations of words/verbs used in our captions.
In our analysis (\cref{tab:statistics}), we follow \cite{goldfarb-tarrant-etal-2020-content} and calculate the percentage of diverse verbs (that are not in the top 5 most frequent verbs) relative to all verbs. Following~\cite{shetty2017speaking}, we also compute the unique-to-total ratio for word unigrams, bigrams, and trigrams (e.g., the ratio between the number of unique word unigrams to the total number of word unigrams over all captions). We further use the spaCy toolkit~\cite{honnibal2020spacy} to extract and lemmatize verbs and calculate the unique-to-total ratio for \textit{verb} unigrams, bigrams, and trigrams. 
The results in \cref{tab:statistics} show that the captions in the HowToCaption dataset have higher language diversity than ASR subtitles across almost all measures except on verb unigram. We observe that the longer action sequences in HowToCaption are more diverse than ASR subtitles, which demonstrates the high quality of our dataset.

\section{Qualitative Examples}
\label{sec:dataset_examples}

\begin{figure*}[t]

\begin{subfigure}[t]{0.48\linewidth}
        \begin{subfigure}[t]{\linewidth}
                \includegraphics[width=0.49\textwidth]{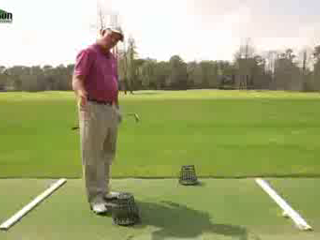}
                \hfill\includegraphics[width=0.49\textwidth]{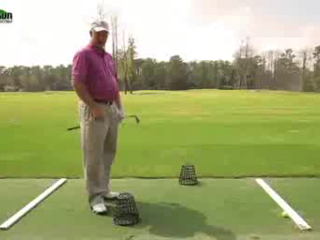}
        \end{subfigure}%
        \caption*{\scriptsize{\textbf{Caption (118s-126s)}: Matt Swanson gives a tip to use buckets to direct the path of the ball \\
        \textcolor{gray}{\textbf{ASR}: 7s: hi i'm matt swanson \\
9s: with matt swanson's we'll go is gonna help me change the direction of your ball play \\
16s: if you're struggling with a slice or a hook and it's moving too much in that direction \\
21s: i'm gonna give you a little tip that we use \\
25s: that's very easy that you can do when you're out to help \\
84s: change that \\
91s: you're slicing now \\
109s: too much \\
116s: so use these buckets when you're out at the range \\
\textbf{120s: move them around to help direct the path} \\
123s: make sure the clubface is closing if you're trying to get rid of the slice opening \\
128s: if you're trying to hit a fade use these tips and you'll get better} \\
        }
        }
\end{subfigure}%
\hfill
\begin{subfigure}[t]{0.48\linewidth}
        \begin{subfigure}[t]{\linewidth}
                \includegraphics[width=0.49\textwidth]{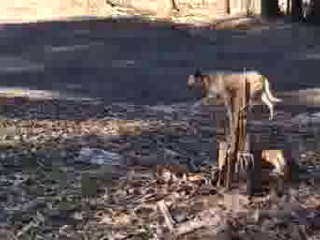}
                \hfill\includegraphics[width=0.49\textwidth]{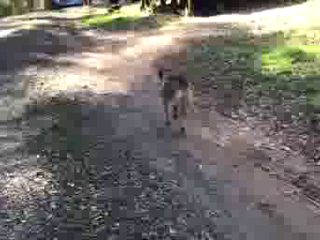}
        \end{subfigure}%
        \caption*{\scriptsize{\textbf{Caption (87s-95s)}: Dog wants to hang out near dirt or other dogs with bones to acquire more bones\\
        \textcolor{gray}{\textbf{ASR}: \textbf{87s: so this is stage one of hiding the bone}\\
90s: burying the bone\\
90s: there's so much more involved\\
92s: let's watch how she behaves\\
94s: when she comes back from burying the bone she becomes a bit annoying\\
97s: she wants to hang out next to dirt or somebody else who has a bone so that she might acquire yet another bone\\
103s: now what i have found in my experience with her and remember every dog is different\\
107s: nobody does things the same way but she lets those bones ferment for about two days before she goes back and finally retrieves the bone\\
115s: and then she sits down and enjoys it\\
116s: and then something makes it really special\\
...
}}}
\end{subfigure}%

\begin{subfigure}[t]{0.48\linewidth}
        \begin{subfigure}[t]{\linewidth}
                \includegraphics[width=0.49\textwidth]{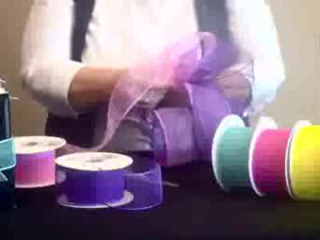}
                \hfill\includegraphics[width=0.49\textwidth]{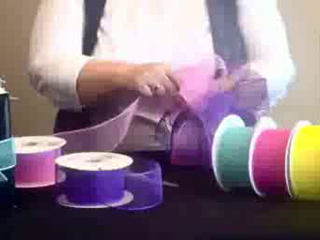}
        \end{subfigure}%
        \caption*{\scriptsize{\textbf{Caption (187s-195s)}: Making a bow with two colors \\
        \textcolor{gray}{\textbf{ASR}: 185s: so if you don't get your your ribbon twisted there's no up or down side to it\\
\textbf{191s: so it's not going to really show}\\
193s: now once i get my three loops on my lighter color i'm going to make a little loop and this is basically just to hide my wire\\
202s: that i'm going to use\\
205s: the wire i use either a teen or a 20 gauge wire and what i'm going to do\\
210s: this little loop is going to be my hide for the wire\\
215s: so i just slide that through like that and pull real tight and twist\\
227s: that will keep your loops good and snug once you get done\\
237s: just kind of work your your loops around\\
243s: and now you have a bow that's made with two colors\\
246s: these are great for easter baskets you can use those for mother's day\\
251s: we have an assortment of colors so we even have some for the holidays for fall\\
257s: it's the type that you can use any time of the year and it makes great bows}}}
\end{subfigure}%
\hfill
\begin{subfigure}[t]{0.48\linewidth}
        \begin{subfigure}[t]{\linewidth}
                \includegraphics[width=0.49\textwidth]{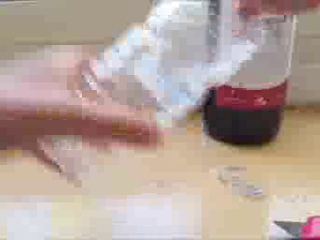}
                \hfill\includegraphics[width=0.49\textwidth]{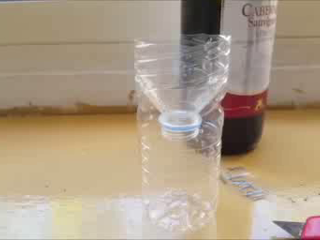}
        \end{subfigure}%
        \caption*{\scriptsize{\textbf{Caption (68s-76s)}: Make sure the bottle stays together  \\
        \textcolor{gray}{\textbf{ASR}: 0s: hi there the amateur scientist here\\
4s: and today in this video i'm going to show you how to make a very cheap and from recycled materials\\
12s: a mosquitoes trap\\
14s: now because we all know that mosquitoes can bite and make you itch and they are very uncomfortable\\
20s: so today in this video i'm going to show you how to make a mosquito trap from a plastic bottle and on their amulet or zoom\\
28s: so let's get into the video\\
30s: first of all we're going to need a plastic bottle and i and some very cheap one\\
45s: you could also use these for sugar\\
61s: so first grab the bottle and and put the upper part into the lower part\\
\textbf{69s: but this yeah and it just stays or it won't get off}\\
\textbf{73s: it's busy here}\\
74s: good deeper cheap wine\\
79s: hello this with it like this or just a little bit more like that\\
...
}}}
\end{subfigure}%

\vspace{-0.25cm}
\caption{ \fix{\small{\textbf{Extended example of video-captions pairs from our \datasetname\ dataset (an extension of Fig. 3 of the main paper)}. The ASR subtitles within the corresponding video clip are bolded. We note that some details in the generated captions are derived from a long ASR context. \label{fig:ext_dataset}}}}
\vspace{-0.25cm}

\end{figure*}

\begin{figure*}[]

\begin{subfigure}[t]{0.48\linewidth}
        \begin{subfigure}[t]{\linewidth}
                \includegraphics[width=0.49\textwidth]{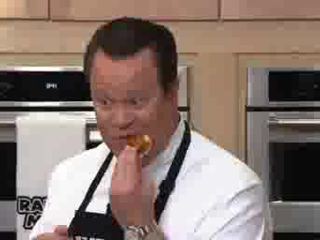}
                \hfill\includegraphics[width=0.49\textwidth]{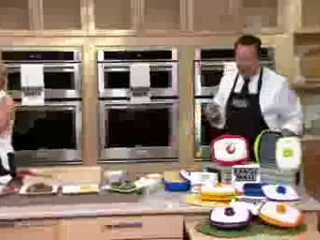}
        \end{subfigure}%
        \caption*{\scriptsize{\textbf{Caption}: Video segment starts with a shot of David's face, which is described as funny \\
        \textcolor{gray}{\textbf{ASR}: and the bottom is actually has holes in it because it gets so incredibly hot so you cannot submerge it in water so we ask you to just rinse it out real quick look at david's face he is so funny}}}
\end{subfigure}%
\hfill
\begin{subfigure}[t]{0.48\linewidth}
        \begin{subfigure}[t]{\linewidth}
                \includegraphics[width=0.49\textwidth]{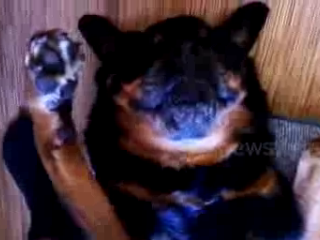}
                \hfill\includegraphics[width=0.49\textwidth]{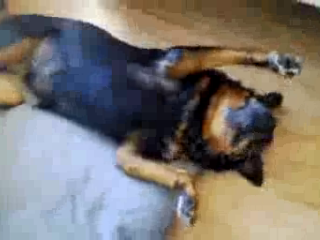}
        \end{subfigure}%
        \caption*{\scriptsize{\textbf{Caption}: Brutus is encouraged to swallow his medication \\
        \textcolor{gray}{\textbf{ASR}: if i put it in a piece of food he'll chew it up and spit it out he knows oh baby i've never seen him do this though}}}
\end{subfigure}%

\begin{subfigure}[t]{0.48\linewidth}
        \begin{subfigure}[t]{\linewidth}
                \includegraphics[width=0.49\textwidth]{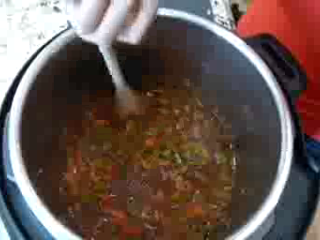}
                \hfill\includegraphics[width=0.49\textwidth]{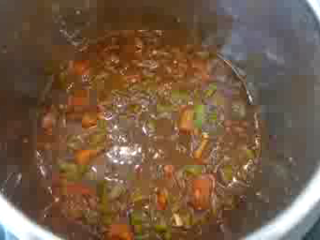}
        \end{subfigure}%
        \caption*{\scriptsize{\textbf{Caption}: Adds two cans of red kidney beans to the chili  \\
        \textcolor{gray}{\textbf{ASR}: you could also use a vegetable broth all right so we're mixing this well}}}
\end{subfigure}%
\hfill
\begin{subfigure}[t]{0.48\linewidth}
        \begin{subfigure}[t]{\linewidth}
                \includegraphics[width=0.49\textwidth]{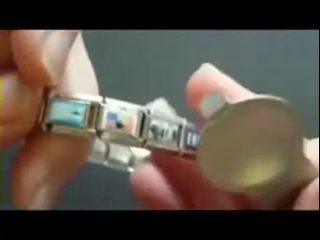}
                \hfill\includegraphics[width=0.49\textwidth]{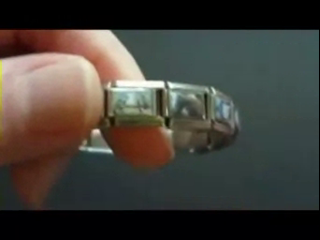}
        \end{subfigure}%
        \caption*{\scriptsize{\textbf{Caption}: She explains the charm tool's little piece of metal acts as a spacer to hold the charms open \\
        \textcolor{gray}{\textbf{ASR}: has this little piece of patootie metal right here that acts as a spacer to hold the charms open so that gap is visible in the back so it's easier to slip the charms on and off }}}
\end{subfigure}%

\begin{subfigure}[t]{0.48\linewidth}
        \begin{subfigure}[t]{\linewidth}
                \includegraphics[width=0.49\textwidth]{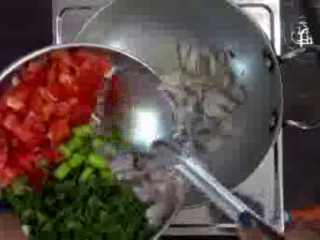}
                \hfill\includegraphics[width=0.49\textwidth]{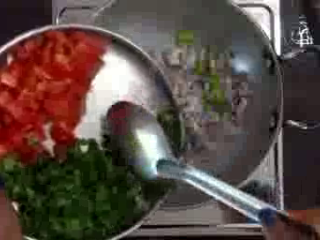}
        \end{subfigure}%
        \caption*{\scriptsize{\textbf{Caption}: Adding chopped onions and green chillies to the pan  \\
        \textcolor{gray}{\textbf{ASR}:once the oil is hot enough we will add our onions and green chillies we need to cook the onions for some time maybe like 2 to 3 minutes until you start noticing that the colors of the onion have changed}}}
\end{subfigure}%
\hfill
\begin{subfigure}[t]{0.48\linewidth}
        \begin{subfigure}[t]{\linewidth}
                \includegraphics[width=0.49\textwidth]{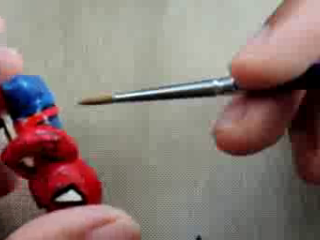}
                \hfill\includegraphics[width=0.49\textwidth]{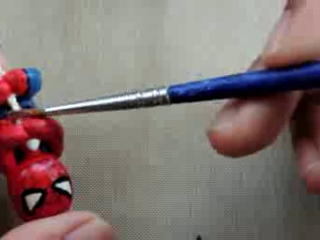}
        \end{subfigure}%
        \caption*{\scriptsize{\textbf{Caption}: Paints the top part \\
        \textcolor{gray}{\textbf{ASR}: i also notice how the blue continues onto the front of him just like right there so be careful with that next you take the white color and you would paint the webbing that he's hanging from here and also his eyes}}}
\end{subfigure}%

\begin{subfigure}[t]{0.48\linewidth}
        \begin{subfigure}[t]{\linewidth}
                \includegraphics[width=0.49\textwidth]{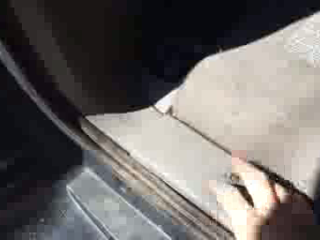}
                \hfill\includegraphics[width=0.49\textwidth]{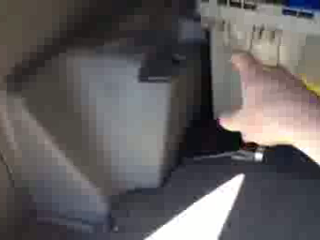}
        \end{subfigure}%
        \caption*{\scriptsize{\textbf{Caption}: Shows where wire runs along inside of vehicle  \\
        \textcolor{gray}{\textbf{ASR}: then ran alongside the gasket right here and runs down here and then this we took off and then ran the wiring in through here put this back down}}}
\end{subfigure}%
\hfill
\begin{subfigure}[t]{0.48\linewidth}
        \begin{subfigure}[t]{\linewidth}
                \includegraphics[width=0.49\textwidth]{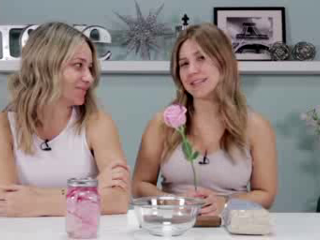}
                \hfill\includegraphics[width=0.49\textwidth]{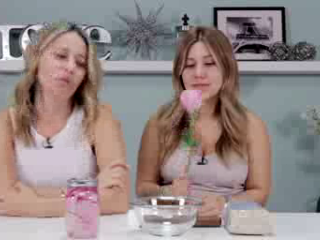}
        \end{subfigure}%
        \caption*{\scriptsize{\textbf{Caption}: You are making a rose petal exfoliating face scrub \\
        \textcolor{gray}{\textbf{ASR}: you guys one of our favorite diys ever had to do with rose petals so we thought let's make another one}}}
        
\end{subfigure}%

\caption{ \small{\textbf{Examples video-captions pairs from our \datasetname\ dataset.} Since ASR subtitles' timestamps do not always correspond to the timestamps of video clips from the \datasetname\ dataset, we show ASR subtitles that intersect with video clip boundaries.  \label{fig:dataset2}}}

\end{figure*}

\begin{figure*}[]

\begin{subfigure}[t]{0.48\linewidth}
        \begin{subfigure}[t]{\linewidth}
                \includegraphics[width=0.49\textwidth]{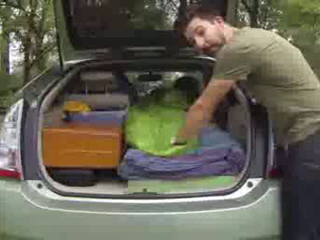}
                \hfill\includegraphics[width=0.49\textwidth]{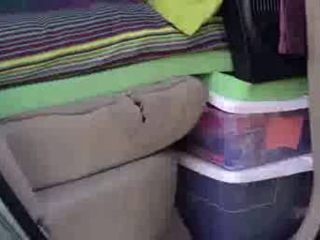}
        \end{subfigure}%
        \caption*{\scriptsize{\textbf{Caption}: Soft bed in the car \\
        \textcolor{gray}{\textbf{ASR}: but honestly when people see a lone prius in the parking lot no one thinks hey i wonder if someone's sleeping in there because come on it's a prius what fits in my car i have a soft bed i have a closet blackout curtains a desk kitchen table and chair a pantry a bike a laundry basket travel kit for emergencies}}}
\end{subfigure}%
\hfill
\begin{subfigure}[t]{0.48\linewidth}
        \begin{subfigure}[t]{\linewidth}
                \includegraphics[width=0.49\textwidth]{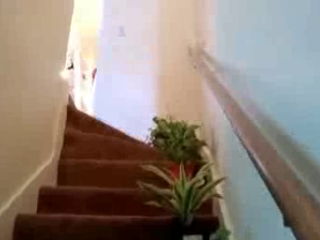}
                \hfill\includegraphics[width=0.49\textwidth]{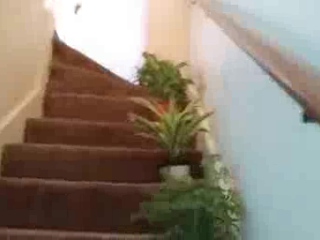}
        \end{subfigure}%
        \caption*{\scriptsize{\textbf{Caption}: Walk upstairs to show light in the ceiling \\
        \textcolor{gray}{\textbf{ASR}: i placed them over here because it's a little bit lighter on this side of the stair case then the other side there's only one light in the ceiling here so i'm gonna walk upstairs and i'm gonna let you see it from the top of the stairs one more time}}}
\end{subfigure}%

\begin{subfigure}[t]{0.48\linewidth}
        \begin{subfigure}[t]{\linewidth}
                \includegraphics[width=0.49\textwidth]{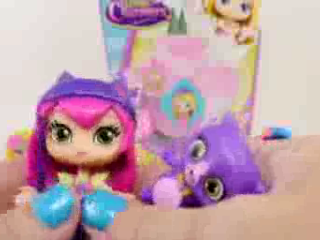}
                \hfill\includegraphics[width=0.49\textwidth]{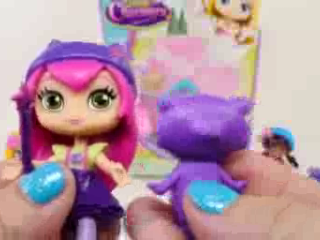}
        \end{subfigure}%
        \caption*{\scriptsize{\textbf{Caption}: Asks viewers to choose favorite pet  \\
        \textcolor{gray}{\textbf{ASR}: i think i like them all for different reasons so that's hard so they can all three be my favorite can't they and look at her little friend }}}
\end{subfigure}%
\hfill
\begin{subfigure}[t]{0.48\linewidth}
        \begin{subfigure}[t]{\linewidth}
                \includegraphics[width=0.49\textwidth]{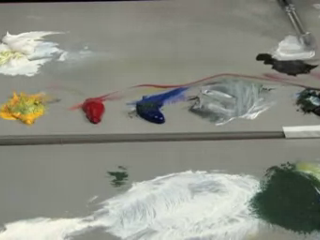}
                \hfill\includegraphics[width=0.49\textwidth]{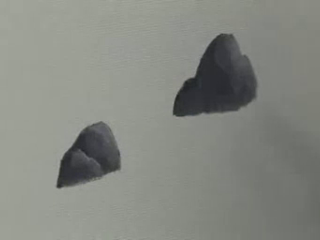}
        \end{subfigure}%
        \caption*{\scriptsize{\textbf{Caption}: The speaker adds white paint to the brush to keep the color bright  \\
        \textcolor{gray}{\textbf{ASR}: so i lay it on with the flat of the brush which deposits it a little heavier it holds up a little better and notice i keep adding white as ...
        }}}
\end{subfigure}%

\begin{subfigure}[t]{0.48\linewidth}
        \begin{subfigure}[t]{\linewidth}
                \includegraphics[width=0.49\textwidth]{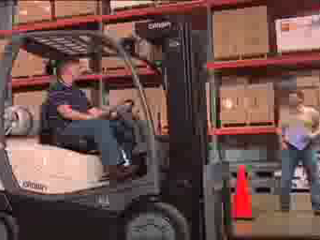}
                \hfill\includegraphics[width=0.49\textwidth]{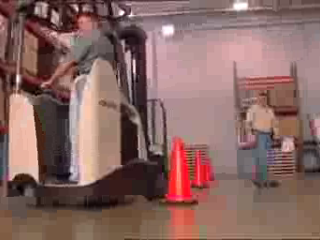}
        \end{subfigure}%
        \caption*{\scriptsize{\textbf{Caption}: DP Move Safe lets operators get out of the classroom and out of their truck faster where they learn how to perform every task and do it safely \colorbox{pink}{(failure)}  \\
        \textcolor{gray}{\textbf{ASR}: dp move safe lets operators get out of the classroom and out of their truck faster where they learn how to perform every task and do it safely }}}
\end{subfigure}%
\hfill
\begin{subfigure}[t]{0.48\linewidth}
        \begin{subfigure}[t]{\linewidth}
                \includegraphics[width=0.49\textwidth]{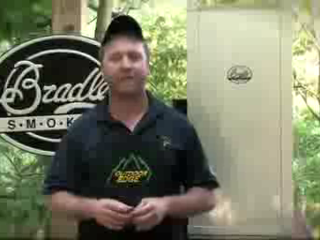}
                \hfill\includegraphics[width=0.49\textwidth]{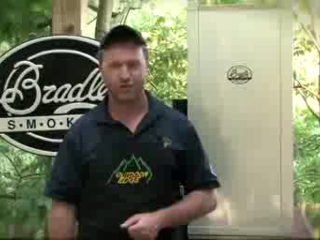}
        \end{subfigure}%
        \caption*{\scriptsize{\textbf{Caption}: Outdoor Edge has instructional gated processing DVDs available on their website \colorbox{pink}{(failure)} \\
        \textcolor{gray}{\textbf{ASR}: this is one of the big issues with large diameter sausage products remember processing your own wild game animal can be fun easy and very rewarding if you have the tools and the knowledge to do the job you're watching outdoor edges}}}
\end{subfigure}%

\begin{subfigure}[t]{0.48\linewidth}
        \begin{subfigure}[t]{\linewidth}
                \includegraphics[width=0.49\textwidth]{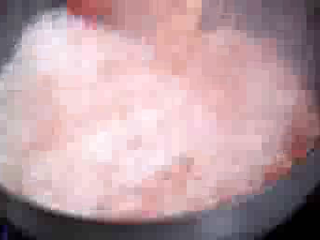}
                \hfill\includegraphics[width=0.49\textwidth]{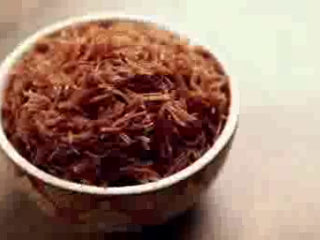}
        \end{subfigure}%
        \caption*{\scriptsize{\textbf{Caption}: Cover it with a lid for 15 minutes until the seviayan is cooked  \colorbox{pink}{(failure)}  \\
        \textcolor{gray}{\textbf{ASR}: today, we will prepare a sweet recipe.. ..called 'sevaiyan' (vermicelli) so come on, let's see how to make sweet sevaiyan.}}}
\end{subfigure}%
\hfill
\begin{subfigure}[t]{0.48\linewidth}
        \begin{subfigure}[t]{\linewidth}
                \includegraphics[width=0.49\textwidth]{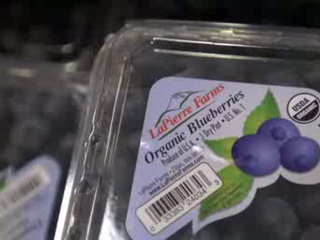}
                \hfill\includegraphics[width=0.49\textwidth]{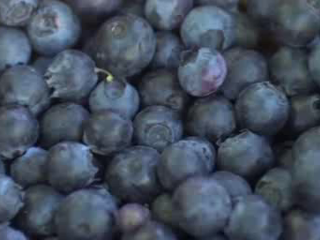}
        \end{subfigure}%
        \caption*{\scriptsize{\textbf{Caption}: The group harvests fresh berries from the farm \colorbox{pink}{(failure)} \\
        \textcolor{gray}{\textbf{ASR}: the kids and i are here at a local blue red patches to manage to harvest some fresh berries at pcc are fresh and frozen organic  ...
        }}}
\end{subfigure}%

\caption{ \small{\textbf{Examples video-captions pairs from our \datasetname\ dataset.} Failure cases are marked as \colorbox{pink}{(failure)}. Since ASR subtitles' timestamps do not always correspond to the timestamps of video clips from the \datasetname\ dataset, we show ASR subtitles that intersect with video clip boundaries.  \label{fig:dataset3}}}

\end{figure*}

\myparagraph{HowToCaption Dataset.} \fix{We present an extension of Fig. 3 from the main paper with video-text examples of our HowToCaption dataset and corresponding ASR subtitles in \cref{fig:ext_dataset}. We see that our HowToCaption method effectively transforms noisy ASR subtitles into proper captions, leveraging the complete ASR context for caption generation. } We demonstrate additional video-text examples of our HowToCaption dataset in~\cref{fig:dataset2} and \cref{fig:dataset3}.  In~\cref{fig:dataset3}, we also showcase instances of failure cases. One such case involves a failure where the LLM was unable to generate a caption and instead copied the input ASR subtitles: ``DP Move Safe lets operators get out of the classroom...'' However, in this example, the ASR subtitles contain a third-person video description with a subjet+verb+object sentence structure that justifies the coping input description without modification. Other failure cases include video-caption pairs, where the caption corresponds to the video only partially, e.g., ``Cover it with lid'' action is not visible on the video while ``until the seviayan is cooked'' is visible.

\myparagraph{LLM Caption Generation.} 
In~\cref{tab:LLM_output_examples}, we showcase captions generated by the Vicuna-13B model, presenting both the input ASR subtitles and their corresponding generated captions for comparison.
We observe the LLM is able to transform scrambled ASR subtitles into ``human-written-like'' descriptions. However, we also note that sometimes LLM fails to produce descriptions. We present some failure cases in~\cref{tab:LLM_output_failures}, which include 1) direct input repetition: instances where the LLM duplicates ASR input without modification; 2) ineffective reformulation: the LLM attempts to convert ASR content into descriptions using ineffective structures like "A person says..."; 3) failure to follow the requested structure: instances where the LLM output doesn't follow ``a timestamp: a sentence'' structure for output, e.g., using ``Summary: '' to write a video description without timestamps.


\begin{table*}[t]
\setlength{\tabcolsep}{3pt}
    \caption{\small{
    \textbf{Examples of caption generation with the Vicuna-13B.} Input ASR subtitles and output generated captions by the Vicuna-13B (without any post-processing).  For brevity, lengthy examples are shortened with the symbol ``...''.  
    \label{tab:LLM_output_examples}
}}
\vspace{-0.25cm}
   \centering
    \resizebox{1\linewidth}{!}{
    \footnotesize
    \begin{tabular}{p{8cm}|p{8cm}} %
\toprule
 Input ASR subtitles & Generated captions \\
\midrule 
0s: hi guys it is bill with septic flow

4s: i am here at a brand new construction that i'm actually working on

8s: here

9s: on the back is the septic field

10s: we've already had bulldozing contractor actually cover up the the septic system

15s: but i'm going to show you why

17s: you want to use septic flow to pre perk the system

22s: septic flow is a product that actually helps get rid of sodium and phosphate buildup and we're actually going to just dump it right here in this two - inch

29s: it goes right out there

29s: we're going to run some water behind it for new construction

33s: the reason you want to do that is because we are actually pre perking the system getting it ready to take those phosphates and sodium buildup

41s: that happens through all different kinds of things

43s: you know whether you're using

44s: i've had people ask me can you use organic detergent

47s: will that prevent sodium buildup

50s: absolutely not

50s: soap by nature of the saponification process that it goes through it's just part of it & 

0s: Bill is at a new construction site.

4s: He is showing the septic field.

8s: The bulldozing contractor has covered up the septic system.

10s: Bill explains why they use septic flow.

17s: Septic flow helps get rid of sodium and phosphate buildup.

22s: Bill demonstrates how they will dump septic flow in a two-inch area.

29s: They will run water behind it for new construction.

33s: Bill explains that they are pre-perking the system.

41s: The system needs to be prepared to take away phosphates and sodium buildup.

44s: People ask if using organic detergents can prevent sodium buildup.

50s: The answer is no, soap is part of the saponification process and will cause buildup.

 \\ %
\midrule

104s: the great thing about the smart chicken

107s: with it being air chilled

108s: it's going to pick up a lot of these flavors when we put the the ginger the garlic

111s: and the sage in here

112s: it's just going to combine well with the meat especially while it's cooking on the grill

116s: all those flavors are just going to pop once the sandwich is complete and once it's all done

121s: now that we have it all mixed i'm just going to separate that out into two patties

129s: um you don't have to pat it up too much

131s: it takes a lot to get it on the grill so just kind of separate it out into two balls

136s: even

136s: and that's our chicken

138s: it's ready for the grill

... &

104s: The speaker prepares the chicken by air-chilling it and adding ginger, garlic, and sage.

108s: The chicken will pick up the flavors while cooking on the grill.

111s: The speaker mixes the chicken and its seasoning.

116s: The flavors will pop once the sandwich is complete.

121s: The speaker separates the chicken mixture into two patties.

129s: The patties are not patted too much, as they will be grilled soon.

131s: The chicken is ready for the grill.

138s: The grill should be hot, around 400 degrees, for grilling the chicken.

141s: The chicken will flatten itself out and get a nice sear on the bottom when grilled.

149s: The speaker recommends not touching the chicken for about four minutes so the bottom can set up.

157s: The chicken will fall apart if touched too soon.  \\ %

\midrule

9s: hello home bruce and welcome to my channel

15s: so i am down to my last bottle of strawberry wine

17s: i've got some in a glass so i thought it would be a fantastic idea to make some more

23s: so it is a bit early in the season to be getting strawberries

27s: it has been rather cold so i am actually using frozen strawberries so i hope you enjoyed the video and i'm gonna have a drink

42s: so for this recipe i am actually going to be using a beauty bucket

46s: in my case it is an up cycled ten liter mayonnaise dough which you would have seen in the sakae video

52s: now if you don't have a brewing bucket of some description which has got a wide mouth so it's less likely to puke

... &

9s: Hello and welcome to the channel.

15s: The host is down to their last bottle of strawberry wine.

17s: They have some wine in a glass and decide to make more.

23s: The host mentions that it's early in the season for strawberries.

27s: The host is using frozen strawberries in the recipe.

42s: The host is using a beauty bucket for the recipe.

46s: The bucket is an upcycled mayo container.

52s: The host mentions that a wide-mouthed bucket is important to avoid puking.

59s: The host mentions that puking is a common issue in fruit brewing.

62s: The host mentions a strawberry brewing bucket that they have.

... \\

\arrayrulecolor{black}\bottomrule
\end{tabular}
}

\end{table*} 
\begin{table*}[t]
\setlength{\tabcolsep}{3pt}
\caption{\small{
    \textbf{Illustrative \colorbox{pink}{failures} in caption generation with the Vicuna-13B.} Input ASR subtitles and output generated captions where the Vicuna-13B failed to generate video descriptions based on the subtitles. Failures include 1) direct input repetition; 2) ineffective reformulation of the ASR subtitles into descriptions using structures like ``A person says .. ''; 3) failure to follow the requested structure ``a timestamp: a sentence''. For brevity, lengthy examples are shortened with the symbol ``...''. 
    \label{tab:LLM_output_failures}
}}
\vspace{-0.25cm}
   \centering
    \resizebox{1\linewidth}{!}{
    \footnotesize
    \begin{tabular}{p{8cm}|p{8cm}} %
\toprule

\textbf{Input ASR subtitles} & \textbf{LLM output: generated captions} \\

\midrule

 2s: i got my barbecue shoes on

9s: go down to the house the sausage

14s: i got my barbecue shoes

16s: welcome to the barbecue pit boys

18s: hi

18s: and today we're gonna do some country meatballs and gravy on the grill

22s: all right

23s: it's real easy to get started here

25s: we got some large granulated salt we got some black peppercorn there we have some granulated onion here

31s: we got some red pepper flakes and here we got some granulated garlic

36s: now instead of using those spices you could just say use your favorite blend of seasonings here

... & 

2s: i got my barbecue shoes on

9s: go down to the house the sausage

14s: i got my barbecue shoes

16s: welcome to the barbecue pit boys

18s: hi

18s: and today we're gonna do some country meatballs and gravy on the grill

22s: all right

23s: it's real easy to get started here

25s: we got some large granulated salt we got some black peppercorn there we have some granulated onion here

31s: we got some red pepper flakes and here we got some granulated garlic

36s: now instead of using those spices you could just say use your favorite blend of seasonings here

... \\

\midrule 
426s: you need three out of five

432s: so first jack jack

435s: ready

436s: oh good man

437s: oh no

438s: come on

448s: everything

449s: here we go

449s: okay actually okay

450s: and then for -

451s: since you're a little older you guys stand behind the line here

456s: so let's have you stand behind jack jack's face

462s: so stand up

463s: okay here we go back there

... &

426s: A person announces that three out of five is needed.

432s: Jack Jack is mentioned.

435s: Someone says "ready."

436s: A person says "oh good man."

437s: A person says "oh no."

438s: Someone says "come on."

448s: Everything is mentioned.

449s: The current video segment starts.

450s: The game is continued.

451s: A person tells the players to stand behind the line.

456s: Jack Jack stands behind his face.

462s: The game continues.

463s: Someone tells Jack Jack to stand up.

465s: The game continues behind Jack Jack's face.

471s: Three is mentioned.

... \\ %
\midrule

3s: so we got to the campground

7s: we want to use the oven

10s: we need to light the manual pilot

11s: we only have to light it once for the whole weekend if we're not moving around

15s: so all we need to do is make sure the propane works the oven works or the stovetop works

22s: now we just need to use the oven

24s: so now we're down at the ovens level

26s: we're going to light the pilot light so all we need to do is have a stick lighter

35s: turn the knob to pilot push and hold light

41s: the pilot the pilots burning one two three i let go didn't turn the knob the pilots staying lit

49s: the pilot is good for all weekend

51s: when you go to cook your pizza set it to the temperature you want pizzas done turn it back to pilot and the pilot will stay lit all weekend

63s: when you're ready to leave then we turn it to off that turns off the pilot

69s: we don't want that pilot on while we're driving all the time and constantly using propane

75s: that's why they give us the option

80s: pilot or off is off &

3s: Campground

7s: Oven

10s: Manual pilot light

11s: Light pilot once for the weekend

15s: Check if propane and oven work

22s: Use the oven

24s: At oven's level

26s: Light pilot

35s: Turn knob to pilot, push and hold

41s: Pilot lit

49s: Good for all weekend

51s: Set temperature, turn to pilot, and leave

63s: Turn off when ready to leave

69s: Off while driving

75s: Option to choose pilot or off

80s: Off is off.

Summary: A group checks if the campground's oven works and lights the manual pilot. They only need to light it once for the weekend as long as they're not moving around. They use the oven and make sure the propane works before they leave. The option to choose pilot or off is given. The pilot should be off while driving.

 \\ %

\arrayrulecolor{black}\bottomrule
\end{tabular}
}

\end{table*} 

\section{Limitations}  
\label{sec:limitations}

\fix{
Our method relies on pre-trained large foundational models, including the large language model Vicuna~\cite{vicuna2023} and the vision-language model BLIP~\cite{li2022blip}. Consequently, our HowToCaption method and the proposed HowToCaption dataset may inherit limitations present in these models. Notably, large language models have several shortcomings, such as biases from their training data, which can lead to the generation of potentially misleading content and the propagation of societal biases~\cite{ray2023chatgpt}. Additionally, since our text-video model is initialized from the BLIP model that was pre-trained on curated and filtered datasets, it might be less robust to noisy low-quality videos in our alignment \& filtering step. In our robustness analysis, we found that the model is capable of filtering noisy input, but 1-2\% of noisy data is still passing the filter. Finally, since our dataset is sourced from the HowTo100M~\cite{miech2019howto100m} dataset, it follows the same data distribution, focusing solely on ``how-to'' topics. While we show improvement across many different tasks, it may limit its applicability to certain downstream tasks.
}

\section{Responsibility to Human Subjects}
\label{sec:responsibility}

\fix{Our HowToCaption dataset is sourced from the publicly accessible HowTo100M dataset~\cite{miech2019howto100m}, which was collected from YouTube. The video content consists of user uploads and is publicly available. However, since the dataset provides only video IDs for download, users who opt out of YouTube are consequently excluded from the HowTo100M and HowToCaption datasets.
We are not aware whether consent was obtained from the users to be included in the original dataset. The dataset may include celebrities or other YouTube-famous individuals. 
}

\end{document}